\documentclass[10pt,twocolumn,letterpaper]{article}

\usepackage{iccv}
\usepackage{times}
\usepackage{epsfig}
\usepackage{graphicx}
\usepackage{amsmath}
\usepackage{amssymb}
\usepackage{multicol}
\usepackage{wrapfig}
\usepackage{lscape}
\usepackage{caption}
\usepackage{booktabs}
\usepackage{tabularx}
\usepackage{array}

\usepackage{algorithm2e}

\usepackage{bm}
\usepackage{bbm}

\usepackage{color}
\usepackage{xcolor}

\usepackage[pagebackref=true,breaklinks=true,letterpaper=true,colorlinks,bookmarks=false]{hyperref}

\usepackage{soul}

\input{commands}

\definecolor{snsgray}{RGB}{179, 179, 179}
\definecolor{snsorange}{RGB}{252, 141, 98}
\definecolor{snsblue}{RGB}{141, 160, 203}

\definecolor{coolgrey}{RGB}{157,157,157}
\definecolor{lightgrey}{RGB}{235,238,238}
\definecolor{lightteal}{RGB}{198,211,222}
\definecolor{cyan}{RGB}{136, 204, 238}
\definecolor{teal}{RGB}{68, 170, 153}
\definecolor{sand}{RGB}{221, 204, 119}
\definecolor{rose}{RGB}{204, 102, 119}
\definecolor{red}{RGB}{250, 94, 91}
\definecolor{orange}{RGB}{255, 200, 63}
\definecolor{yellow}{RGB}{254, 239, 109}

\definecolor{darkgreen}{rgb}{0.09, 0.45, 0.27}

\usepackage{xcolor}
\definecolor{codegreen}{rgb}{0,0.6,0}
\definecolor{codegray}{rgb}{0.5,0.5,0.5}
\definecolor{codepink}{RGB}{252, 142, 172}
\definecolor{codepurple}{rgb}{0.58,0,0.82}
\definecolor{backcolour}{RGB}{245,245,245}
\usepackage{listings}
\lstdefinestyle{mystyle}{
    backgroundcolor=\color{backcolour},   
    commentstyle=\color{magenta},
    keywordstyle=\color{blue},
    numberstyle=\tiny\color{codegray},
    stringstyle=\color{codepurple},
    basicstyle=\fontfamily{\ttdefault}\footnotesize,
    breakatwhitespace=false,         
    breaklines=true,                 
    captionpos=b,                    
    keepspaces=true,    
    frame=single,
    numbersep=5pt,                  
    showspaces=false,                
    showstringspaces=false,
    showtabs=false,                  
    tabsize=2,
    classoffset=1, %
    otherkeywords={range},
    keywordstyle=\color{violet},
    classoffset=0,
}

\iccvfinalcopy %

\def\iccvPaperID{****} %

\begin{document}

\title{When Learning Is Out of Reach, Reset: Generalization in Autonomous Visuomotor Reinforcement Learning}

\author{
\\[-0.3in]\textbf{Zichen Zhang$^\dagger$, Luca Weihs$^\dagger$}\\
$^\dagger$PRIOR @ Allen Institute for AI 
\\ \href{https://zcczhang.github.io/rmrl}{\texttt{https://zcczhang.github.io/rmrl}}
}

\ificcvfinal\thispagestyle{empty}\fi

\twocolumn[{%
\renewcommand\twocolumn[1][]{#1}%
\maketitle
\begin{center}
    \centering
    \captionsetup{type=figure}
    \vspace{-1em}
    \includegraphics[width=0.9\textwidth]{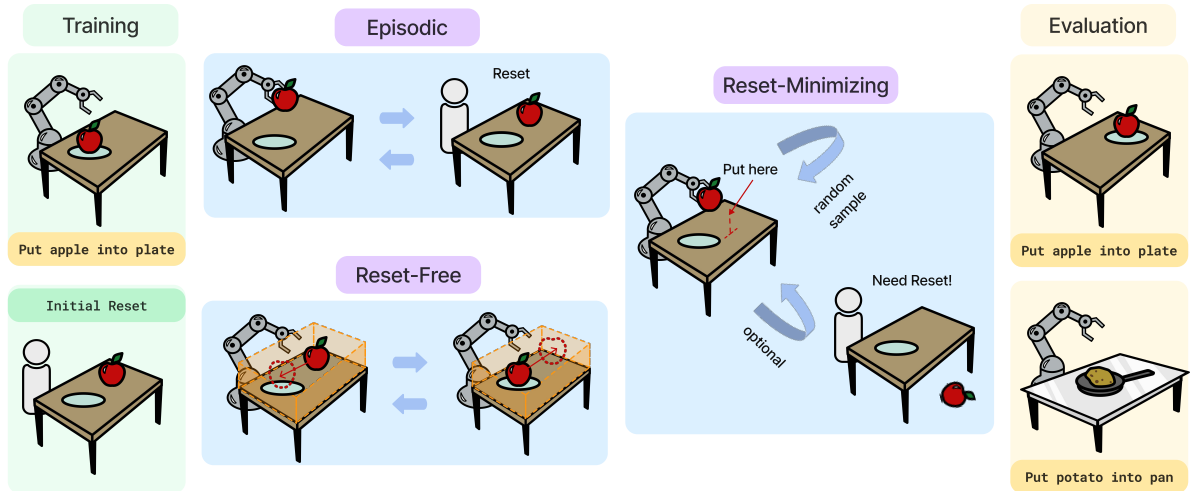}
    \captionof{figure}{\textbf{Episodic, Reset-Free, and Reset-Minimizing RL.} 
    In standard (\ie episodic) reinforcement learning (RL) agents have their environments reset after every success or failure, an expensive operation in the real world. In Reset-Free RL (RF-RL), researchers have designed ``reset games'' which allow for learning so long as special care is taken to avoid irreversible transitions (\eg an apple falling out of reach). We consider Reset-Minimizing RL (RM-RL) where in realistic and dynamic environments agents may request human interventions but should minimize these requests. 
    }\label{fig:teaser}
\end{center}%
}]

\begin{abstract}
    \vspace{-1em}
   Episodic training, where an agent's environment is reset to some initial condition after every success or failure, is the de facto standard when training embodied reinforcement learning (RL) agents. The underlying assumption that the environment can be easily reset is limiting both practically, as resets generally require human effort in the real world and can be computationally expensive in simulation, and philosophically, as we'd expect intelligent agents to be able to continuously learn without external intervention. Work in learning without any resets, i.e{.} Reset-Free RL (RF-RL), is very promising but is plagued by the problem of irreversible transitions (e.g{.} an object breaking or falling out of reach) which halt learning. Moreover, the limited state diversity and instrument setup encountered during RF-RL means that works studying RF-RL largely do not require their models to generalize to new environments. 
   In this work, we instead look to minimize, rather than completely eliminate, resets while building visual agents that can meaningfully generalize. As studying generalization has previously not been a focus of benchmarks designed for RF-RL, we propose a new Stretch Pick-and-Place (\bench) benchmark
   designed for evaluating generalizations across goals, cosmetic variations, and structural changes.
   Moreover, towards building performant reset-minimizing RL agents, we propose unsupervised metrics to detect irreversible transitions and a single-policy training mechanism to enable generalization.
   Our proposed approach significantly outperforms prior episodic, reset-free, and reset-minimizing approaches achieving higher success rates with fewer resets in \bench and another popular RF-RL benchmark. Finally, we find that our proposed approach can dramatically reduce the number of resets required for training other embodied tasks, in particular for RoboTHOR ObjectNav we obtain higher success rates than episodic approaches using 99.97\% fewer resets.

\end{abstract}

\section{Introduction}\label{sec:intro}

A common assumption made when training embodied agents using reinforcement learning (RL) is that the agent's environment can be easily reset after every success or failure: \ie, agents are trained \emph{episodically}. For instance, an agent trained to perform visual navigation that finds itself stuck or lost will be helpfully teleported to a new position (or even placed into a new home)~\cite{savva2019habitat,Deitke2020RoboTHORAO,Deitke2022ProcTHOR} and a mobile manipulation agent that throws all objects onto the floor will find those objects back in their original positions when asked to perform its next task~\cite{Ehsani2021Manipulathor,Ehsani2022object,Ni2021towards,Li2021iGibson2O,szot2021habitat}. While this assumption may not induce too high a cost when training agents in simulated environments,\footnote{Even in simulation, resetting can be a computationally expensive operation, especially when done frequently
(see, \eg, ~\cite{Kolve2017AI2THORAI,savva2019habitat,mu2021maniskill,fan2022minedojo,mittal2023orbit}). While rarely documented explicitly, many existing reinforcement learning frameworks for embodied agents (\eg Habitat~\cite{savva2019habitat} and AllenAct~\cite{Weihs2020Allenact}) employ tricks to so as to minimize the cost of resets during training.} 
resetting an agent's environment in the real world can be extraordinarily expensive, frequently requiring human intervention. This combination of a need for frequent human intervention and data-hungry modern reinforcement learning algorithms has inspired researchers, largely within the robotics community, to pursue \emph{Reset-Free Reinforcement Learning} (RF-RL)~\cite{leavenotrace,sharma2021autonomous,zhu2020ingredients,xu2020continual,gupta2021reset,sharma2021earl,xu2022dexterous,sharma2022state,gupta2022demonstration,xie2022ask}. In RF-RL, an agent is placed into an environment in some initial configuration and must, as ``reset-free'' suggests, learn to perform its task without additional external intervention. An ideal RF-RL algorithm would both save significant expense during initial training and would even allow for agents to continually learn during deployment. Unfortunately, we argue that learning in the reset-free setting can be tremendously difficult due to the presence of irreversible transitions, which can halt learning entirely, and limited state diversity, which harms generalization.

First, for an agent to have any hope of learning in a reset-free setting it is critical that the agent avoids all \emph{irreversible transitions}, namely all state transitions that cannot be undone by taking further actions. For instance, a car whose wheel becomes stuck in a pothole has undergone an irreversible transition as it can no longer, without external intervention, escape that pothole. Similarly, a pick-and-place robotic agent that drops an object beyond its grasp can no longer hope to learn how to interact with that object. Such transitions can simply make RF-RL impossible: an agent can't learn without experience. To avoid these irreversible transitions, many existing works frequently either (1) carefully construct their agent's environment so that irreversible transitions are unlikely or impossible (\eg putting boundaries around a table so that a pick-and-place agent cannot drop an object onto the floor)~\cite{gupta2021reset,sharma2021earl,xu2022dexterous}, or (2) provide a set of human demonstrations used to pretrain the agent so that the agent is biased against taking irreversible transitions~\cite{gupta2022demonstration,xie2022ask}. Recent works employ these strategies simultaneously~\cite{Sharma2023SelfImproving}. Both of these approaches have clear disadvantages and are not guaranteed to solved the problem of irreversible transitions. Indeed, even when carefully constructing the environment and using additional supervision, it is a common for works studying RF-RL to not be completely ``reset-free'' as they may reset the environment periodically (albeit after long time horizons)~\cite{sharma2021earl,sharma2022state,xie2022ask} or use heuristics to return at agent to a pre-defined state~\cite{Sharma2023SelfImproving}.

Second, while irreversible transitions can be fatal to reset-free learning, a more pernicious problem is that of \emph{limited state diversity}. In the embodied-AI computer vision community, a large emphasis is placed on an agent's ability to generalize. Indeed, existing popular embodied AI benchmarks require that an agent trained in one set of homes is able to successfully complete its task when placed into unseen testing homes containing unseen object instances~\cite{savva2019habitat,szot2021habitat,Weihs2021Visual,Ehsani2021Manipulathor,Deitke2020RoboTHORAO,Li2021iGibson2O}. When anticipating such significant generalization ability, resets become a critical tool in ensuring that an agent sees a wide variety of unique states. Each reset is an opportunity to place the agent into a room that it has not previously visited or, even, into an entirely new environment. When learning reset-free, however, an agent is constrained to a single environment which it may never exhaustively explore. Moreover, as discussed above, the need to carefully construct environments to avoid irreversible transitions inherently means that the domains in which reset-free agents are trained will be characteristically different than those in which they are deployed. In part due to the above, the problem of reset-free learning has been largely studied in visually simple simulated environments using low-dimensional observations~\cite{sharma2021earl,xie2022ask} or in real-world robotics settings where generalization across environments is not required~\cite{gupta2021reset,xu2022dexterous,Sharma2023SelfImproving}.

We look to study how one may build visual agents which can learn reset-free and can generalize to new environments. Due to the problems of irreversible transitions and limited state diversity, however, we believe that the ``no resets'' requirement is simply too strict. Instead, we propose to study \emph{Reset-Minimizing Reinforcement Learning} (RM-RL) where, during training, an agent can request a reset at any point but must attempt to minimize these requests. Similarly as to how competing computer vision models are compared conditional on their parameter counts, this suggests comparing competing RM-RL algorithms conditional on their reset rates. An RM-RL algorithm that achieves a higher success rate than all competitors that use more resets than itself is optimal for that number of resets.

As prior work in RF-RL 
is primarily set in simple $2D$ or fixed environments where the need for generalization is limited, we create a new \emph{Stretch Pick-and-Place} (\bench) benchmark built in the \thor simulator~\cite{Kolve2017AI2THORAI}. In \bench a mobile manipulation robot (the Stretch RE1\footnote{\url{https://hello-robot.com/product}}) is placed before a table during training and must move a given object to various target locations, specified by a language prompt, on and around the table. During testing, we evaluate the agent's ability to generalize when faced with positional, cosmetic (\eg, colors and materials), and structural (\eg, new object instances and furnished scenes) augmentations.

Towards enabling RM-RL, we make two methodological contributions. As our first, and main, methodological contribution we propose, in Sec.~\ref{sec:quantifying-irr}, a collection of well-motivated unsupervised metrics for characterizing when an agent has experienced an irreversible transition. Using these metrics we show, in our experiments, that we can dramatically improve training efficiency suggesting a general framework where an agent requests resets only when necessary. Then, in Sec.~\ref{sec:single-policy}, we describe how, in contrast to the popular, carefully-designed, forward-backward and task-decomposition methodologies used in RF-RL, using a single policy and a random goal sampling approach results in high training performance and enables generalization.

In summary, our contributions include:
(1) the \bench benchmark for studying reset minimizing reinforcement learning in visual embodied environments, (2) general metrics designed to characterize when agents have undergone (near-)irreversible transitions during autonomous training, (3) a single-policy, random-goal conditioned, learning strategy for autonomous visuomotor RL control, and (4) extensive experimental results across (mobile, continuous) manipulation and navigation domains in which we ablate our proposed approach and show that our learning framework is efficient and enables generalization. Our benchmark and training code are open-sourced at \href{https://zcczhang.github.io/rmrl}{\texttt{https://zcczhang.github.io/rmrl}}.

\section{Related work}

\begin{figure*}[htbp!]
    \centering
    \vspace{-0.3em}
    \includegraphics[width=\textwidth]{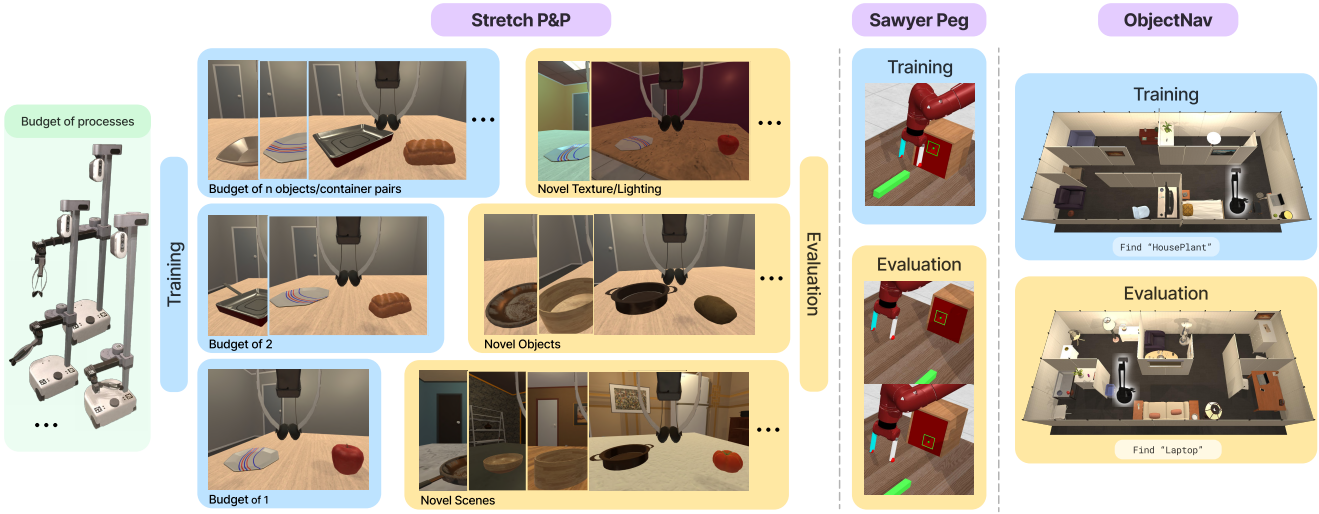}
    \caption{\textbf{Overview of proposed \bench and other experiment environments.} Here we show the training and evaluation configurations for the \bench, Sawyer Peg, and RoboTHOR ObjectNav tasks (from left to right). During training (blue panels), 
    the agent observes: (\bench) as few one household object and one container depending on allowed object budget, (Sawyer Peg) exactly one type of stationary box with a goal hole at its upper center, and (ObjectNav) a limited set of house structures. During evaluation (yellow panels), we require the agent to generalize to: (\bench) novel cosmetic changes, novel object instances, and a combination of the above alongside with other cosmetic and structural background changes, (Sawyer Peg) novel box and hole positions (the hole position is highlight with green here only for visualization purposes), and (ObjectNav) fully unseen house structures.} \label{fig:benchmark}
\end{figure*}

\noindent\textbf{Reset-free learning.}
Reset-free learning has been largely studied in visually simplistic grid and MuJoCo~\cite{Todorov2012Mujoco} style environments~\cite{leavenotrace,xu2020continual,Lu2021LifeLongLearning,sharma2021autonomous,sharma2022state,xie2022ask}, and in well-controlled real-world robotics settings carefully designed to avoid irreversible transitions~\cite{zhu2020ingredients,gupta2021reset,xu2022dexterous,gupta2022demonstration,sharma2021earl,Sharma2023SelfImproving}. One popular approach for reset-free RL introduced by Eysenbach \emph{et~al{.}}~\cite{leavenotrace} (see also~\cite{Han2015LearningCompound, Richter2017SafeVisualNavigation}) involves the joint learning of ``forward'' and ``reset'' (or ``backward'') policies; the forward policy is trained to complete the task of interest (\eg placing a peg into a hole) while the reset policy is trained to reset the environment to some initial state (\eg a peg laying on a table). This forward-reset approach has been further improved, such improvements include the use of curriculum learning to learn harder tasks (VaPRL)~\cite{sharma2021autonomous}, encouraging the development of diverse skills via a discriminator-based approach (LSR)~\cite{xu2020continual}, and the use of small amounts of expert demonstration data to better inform the reset policy as to which distribution of states it should reset (MEDAL)~\cite{sharma2022state}. Unlike these works, we argue that, when attempting to train mobile, generalizable, agents in high-dimensional visual environments, the problems of irreversible transitions and limited state diversity necessitate developing techniques that minimize resets rather than fully eliminating them.

\noindent\textbf{Avoiding irreversible transitions: safe and reversibility-aware RL.} Even in settings where resets are permitted, avoiding irreversible transitions may be preferred for safety or efficiency reasons. Indeed, irreversibility is intimately tied to safety and multiple works have proposed methods to encourage agents to avoid undesirable behaviors via constrained optimization or penalization~\cite{Chow2017RiskConstrained,Achiam2017ConstrainedPolicyOpt,Tessler2019RewardConstrained,Ray2019BenchmarkingSafeExploration,Srinivasan202LearningToBeSafe,Ni2023TowardsDisturbanceFree}. A few other works have studied how explicit knowledge of reversibility can improve agent behavior~\cite{Kruusmaa2007DontDoThings,Grinsztajn2021ThereIsNoTurningBack}. Xie \emph{et al.}~\cite{xie2022ask} consider the problem of irreversible transitions for reset-free RL and propose ``proactive
agent interventions'' (PAINT)~\cite{xie2022ask} a method which extends MEDAL by training a classifier to predict that it has entered irreversible states (using some ground truth labels for such states). This classifier is then used to allow the agent to request a reset when required. PAINT also penalizes agents for entering irreversible states. Unlike this work, we are most concerned with studying the generalization of reset-minimizing agents within high-dimensional embodied environments. Moreover, as we discuss in Section \ref{sec:quantifying-irr}, we argue that ``ground truth'' irreversibility labels frequently suffer from false negatives and so we propose metrics to label such states using unsupervised methods.

\noindent\textbf{Embodied benchmarks.}
In recent years there have been a significant number of benchmarks proposed for the study of training embodied AI models. Among these, we highlight a few related works that study visual navigation~\cite{savva2019habitat,Deitke2020RoboTHORAO,Chen2020Soundspaces,Wani2020Multion,Ramakrishnan2021HabitatMatterport3D,Shen2020iGibsonAS}, visual mobile object manipulation and rearrangement~\cite{szot2021habitat,Li2021iGibson2O,Gan2021Transport,Weihs2021Rearrangement,Ehsani2021Manipulathor,Ehsani2022object}, fine-grained grasping and articulated object manipulation~\cite{mu2021maniskill,mittal2023orbit}, continuous (robotic) control~\cite{Brockman2016Gym,Duan2016Benchmarking,yu2019meta}, and safety~\cite{Ray2019BenchmarkingSafeExploration}. Most related to our work are the Environments For Autonomous Reinforcement Learning (EARL)~\cite{sharma2021earl} and ArmPointNav~\cite{Ehsani2021Manipulathor} benchmarks. The EARL benchmark was designed explicitly to study reset-free learning but, unlike our work, does not emphasize reset-minimization, visual observations, or generalization. The ArmPointNav benchmark is, like our work, a mobile manipulation benchmark set in AI2-THOR but is designed for episodic agent training.

\section{The Stretch Pick-and-Place Benchmark}\label{sec:benchmark}
Existing benchmarks built to study RF-RL and RM-RL focus primarily on visually simplistic environments with low-dimensional state spaces. 
Moreover, these benchmarks are designed to evaluate only in-domain performance: an agent is trained in precisely the same environment, and with the same objects in the same configurations, as in evaluation. As we are interested in studying how RM-RL agents trained with rich, visual, observations are able to generalize when faced with novel objects and visual diversity, we design a new benchmark to evaluate agents in this context.

As we wish to evaluate agents' ability to generalize to both purely cosmetic (\eg, changing the color of materials or lighting of a scene) and structural (\eg, novel objects or scene layouts) changes, we chose to build our benchmark within \thor~\cite{Kolve2017AI2THORAI}, a high visual fidelity simulator of indoor environments. While several other alternative visually rich simulators exist, \eg iGibson 2.0~\cite{Li2021iGibson2O} and Habitat 2.0~\cite{szot2021habitat}, \thor offered both a large set of realistic household object instances and a rich set of tools, see \eg \textsc{ProcTHOR}~\cite{Deitke2022ProcTHOR}, for applying cosmetic and structural augmentations to scenes.

We now define our task. As is highlighted by the struggle of the community to solve even visually simple environments using RF/RM-RL~\cite{sharma2021earl,sharma2022state,xie2022ask}: learning when reset-limited is quite challenging. Given this, an overly complex visual RM-RL benchmark would almost certainly be too difficult to be meaningful in the near term. For this reason, we aim to design a task that, while realistic, does not require excessively long-horizon planning. To this end we design the Stretch Pick\&Place (\bench) task.

\noindent\textbf{Evaluation.} During evaluation in \bench, a Stretch RE1 Robot, see Fig.~\ref{fig:benchmark}, is placed before a table within a room. On this table are two objects, a container (\eg a bowl, plate, \etc) and a small household item (\eg an apple, sponge, \etc). The agent is given a text description of a task involving how the household item should be moved where this instruction can be semantic \eg, ``Put the apple into the plate'', or point-based, ``Put the apple at location $X$'' where $X$ encodes the relative position between the goal coordinate and the agent's gripper. To study generalization, we consider four separate evaluation settings (see Fig.~\ref{fig:benchmark} for visualizations). (1) Positional out-of-domain (\textsc{Pos-OoD}): the environment and objects that must be manipulated are identical to those used during training but the objects' positions and goals are randomized to be much more diverse than those seen in training. 
(2) Visually out-of-domain (\textsc{Vis-OoD}): object instances are the same as in training but the lighting and the materials/colors of background objects will be varied. (3) Novel objects (\textsc{Obj-OoD}): none of the above visual augmentations will be applied but the container and household object instances will be distinct from those seen during training. (4) All out-of-domain (\textsc{All-OoD}): the agent experiences the visual augmentations from (2), novel object instances as in (3), and the addition of new background distractor objects simultaneously. For more details on these evaluation settings, please see Appendix.~\ref{appendix:env:stretch}.

\noindent\textbf{Training.} 
As in evaluation, the agent is placed before a table with a container and a household object. The table, lighting, and object materials are all kept constant during training. We do not place any constraints on the types of tasks that can be used to train the agent, indeed finding good training tasks is a critical area of study for reset-minimizing learning that we wish to encourage. Upon requesting a reset, the agent's position, as well as the position of the two objects, may be placed into any initial configuration. During training, the agent is intentionally limited to a single environment so as to encourage building tools to enable generalization even in this highly constrained setting. As achieving this generalization may be challenging, we do consider allowing more diversity to be introduced during training by allocating a budget of more than one seen object or container (see Fig.~\ref{fig:benchmark}, blue areas for \bench). For instance, with a budget of 2, the agent would be allowed to see 2 household objects and 2 containers during training with one of each object being selected at every reset for the agent to manipulate. We provide full descriptions of simulator metadata, success criteria, and object partitions in Appendix.~\ref{appendix:env:stretch}.

\noindent\textbf{Observations and action space.} As we show in Fig.~\ref{fig:benchmark}, the agent's observation is a 224{$\times$}224 RGB image corresponding to a camera attached to the agent's wrist. We also include proprioceptive sensors corresponding to the agent's arm position. The arm of a Stretch RE1 agent uses a telescoping mechanism to move forward and back, may move up and down, and the gripper has one rotational degree of freedom allowing for changes in yaw, see Figures~\ref{fig:benchmark} and \ref{fig:irreversibility-types} for 3rd person views of the stretch robot. The robotic arm of the Stretch RE1 robot is orthogonal to the agent's forward and backward movement and so, to move the arm laterally, the agent must move its body in the forward and backward direction. To highlight the study of irreversible transitions in our benchmark, and as to not add additional complexity to \bench, we restrict the robot body to not rotate in training, although the wrist of the agent may do so. The maximum rotation for the wrist is 2$^\circ$ per step, and horizontal/vertical arm movement is limited to 5cm per step. With perfect execution, success can generally be achieved within 50 steps, thisis similar to other short-horizon, continuous-space, manipulation tasks~\cite{yu2019meta}. See Appendix.~\ref{appendix:env:stretch} and Table.~\ref{tab:stretch_action} for further details regarding the observation and action spaces.

\section{Methods}\label{sec:methods}

As discussed in Section~\ref{sec:intro}, two fundamental problems when attempting to build generalizable agents in the reset-free setting are irreversible transitions and limited state diversity. As it is often impractical or impossible to guarantee that an agent does not undergo any irreversible transitions during training we will present, in Sec.~\ref{sec:quantifying-irr}, measures that we use to quantify when an agent has undergone such a transition. As we show in our experiments, these measures can be used by the agent to only request resets when it is no longer learning thereby minimizing the total number of resets required to train a performant model. Next, in Sec.~\ref{sec:single-policy}, we take a first step towards building generalizable RM-RL agents; in particular, we propose to do away with the learned forward-reset policies popular in prior RF-RF work and, instead, learn a \textit{single} policy which is presented with randomly generated goals during training. 

\subsection{Quantifying Irreversibility}\label{sec:quantifying-irr}
\begin{figure}[t]
    \centering
    \includegraphics[width=1.0\linewidth]{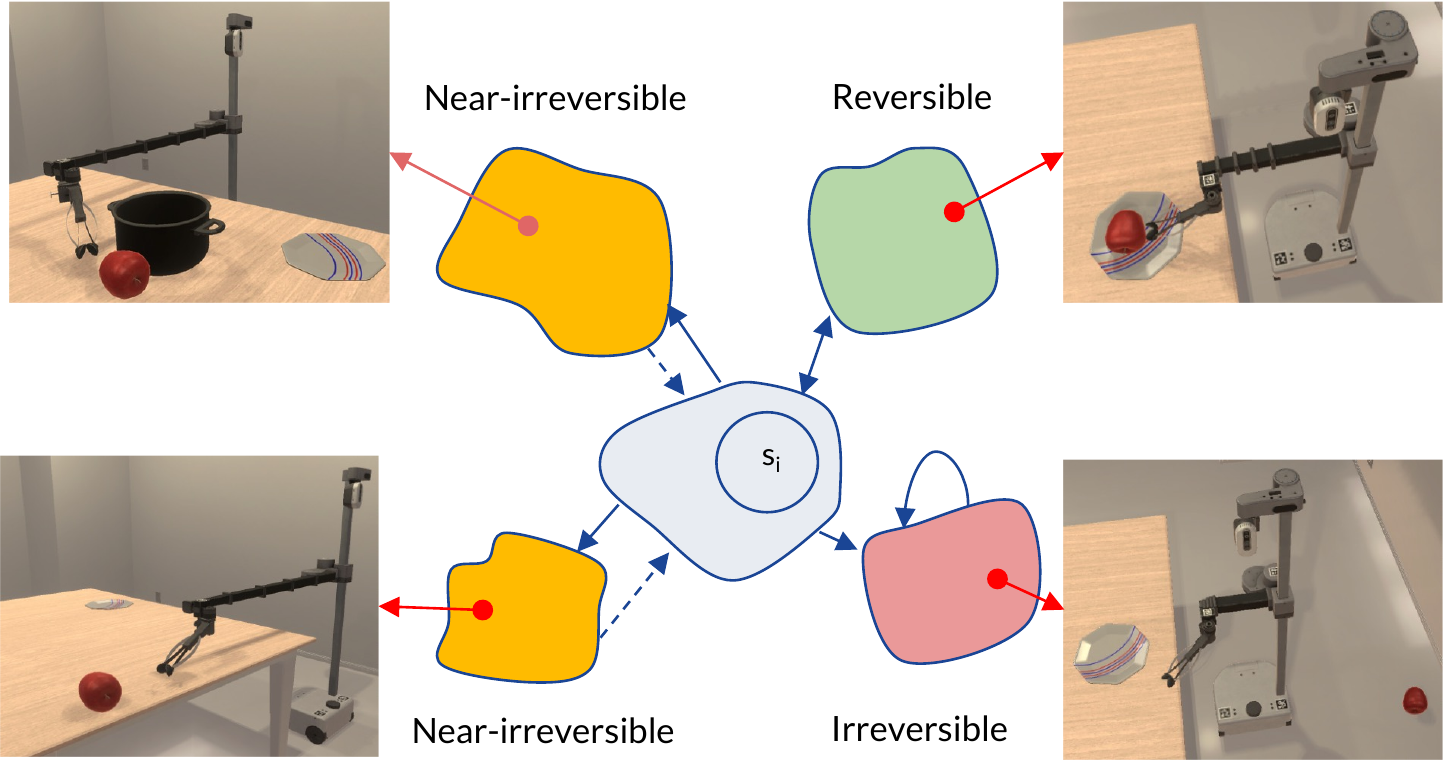}
    \vspace{-1em}
    \caption{\textbf{Reversible and (Near) Irreversible States.} Examples of reversible, irreversible, and NI states for our \bench task. Reversible (top right): the target apple is within easy reaching distance. Irreversible (bottom right): the apple has fallen off the table, as the agent cannot rotate in \bench the apple can no longer be reached. Near-irreversible (left): the apple is in tricky-to-reach locations being behind other objects or at the extreme limits of the arm's reaching capabilities.}\label{fig:irreversibility-types}
    \vspace{-1em}
\end{figure}

\noindent \textbf{Measures of Irreversibility.} Some irreversible transitions are explicit, \eg a glass is dropped and shatters. However, in a more complex real-world environment, they may be more subtle. For instance, when a robot is tasked with cleaning a room, it may encounter situations where some trash is accidentally pushed or blown into hard-to-reach locations, such as under a sofa or in the corners of the room. In such cases, the robot may find it challenging, but not strictly impossible, to pick up or sweep debris back using its regular cleaning tools. We refer to these states that are difficult, but not impossible, to recover from as near-irreversible (NI) states. See Fig.~\ref{fig:irreversibility-types} for examples of reversible, irreversible, and near-irreversible states in our \bench benchmark. Explicitly labeling NI and irreversible states can be challenging, as it is practically infeasible to hard-code all possible types of irreversibilities that may occur in different rooms or with different robots in the real world. Complicating this problem further, the set of near-irreversible states also depends on the agent's policy $\pi$: which states should be considered near-irreversible can, and should, change during training. For instance, an agent near the end of training may be able to recover after pushing trash beneath a sofa but a near-random agent at the start of training would have little hope of doing so. For this reason, we wish to build measures that, by inspecting the recent behavior of the agent, detect when the agent has undergone a (near-)irreversible transition.

As usual in reinforcement learning for embodied agents, we formalize our setting as a Partially Observed Markov Decision Process (POMDP) $(\cS, \cO, \cA, P_T, A, \mu, \gamma)$ with state space $\cS$, partial observation space $\cO$, 
action space $\cA$, transition probability measure $P_T$, reward structure $R:S\times A \rightarrow \mathbb{R}$, initial state distribution $\mu \in \Pi(S)$, and discount factor $\gamma\in(0,1]$. For simplicity of presentation, we will assume that $\cS$ and $\cA$ are discrete. The goal is to learn a policy $\pi$, \ie, a function that maps partial observations to distributions over actions, which maximizes the expected future $\gamma$-discounted
expected return $\mathbb{E}_{\mu, P_T, \pi}\left[\sum_{t=0}^\infty \gamma^t r(s_t, a_t)\right]$. Let $\tau_\pi(i)\in\cS$ for $0\leq i$ be a random variable representing the state of the environment after an agent has executed its policy $\pi$ up to timestep $i$. Here $\tau_\pi(0)$ represents the start state of the agent after a reset (\ie, $\tau_\pi(0)\in \cI$ where $\cI\subset \cS$ is some set of initial starting states). 
For space, we provide more formal definition and analysis of what we mean by a NI transition in Appendix~\ref{appendix:measure}.

Suppose that agent has taken $t$ steps producing the trajectory $\cT_{t}=\{\tau_\pi(0), \ldots, \tau_\pi(t)\}$. Intuitively, undergoing an NI transition should correspond to a decrease in the degrees of freedom available to the agent to manipulate its environment: that is, if an agent underwent an NI transition at timestep $i$ then the diversity of states $\tau_{\pi}(i+1),\ldots, \tau_{\pi}(t)$ should be small compared to the diversity before undergoing the irreversible transition. This intuition suggests that we may define a collection of irreversibility decision functions, \ie functions that return 1 when predicting that the agent has undergone a near-irreversible transition, using a partitioning approach. In particular, we want to ask: for some width $W>0$, how many disjoint, continuous, blocks of width $W$ are there in $\cT_{t}$ where the diversity of states within the block is below some threshold $\alpha>0$. To formalize this, let $P(t) = \{(i_0, i_1, \ldots, i_m): 0=i_0<i_1<...<i_m=t+1,\ m>0\}$. Then we can compute the above count, which we call $\phi_{W,\alpha,d}(\cT_t)$, as \\[-1.5em]
\begin{align*}
    \max_{(i_0, \ldots, i_m)\in P(t)}\sum_{j=0}^{m-1}1_{[i_{j+1}-i_{j} \geq N]}
     \cdot 1_{\{ d\left(\tau_\pi(i_j), \ldots, \tau_\pi(i_{j+1}-1)\right) < \alpha\}}
\end{align*}
~\\[-1em]
where $d:\cS^H\to \bR_{\geq0}$ is some non-negative measure of diversity among $W$ states $s$. As $\phi_{W,\alpha,d}$ is a counting function, we can turn it into a decision function simply by picking some count $N>0$ and deciding to reset when $\phi_{W,\alpha,d}\geq N$. In particular, we will let $\Phi_{W,N,\alpha,d}$ be the function that equals 1 if and only if $\phi_{W,\alpha,d}\geq N$. In our experiments we evaluate several diversity measures $d(s_1,\ldots, s_H)$ including: (1) an empirical measure of entropy, (2) the mean standard deviation of the $s_i$, and (3) a euclidean distance-based measure, and (4) a distance measure using dynamic time warping (DTW). Details and algorithm pseudocode of these measures are provided in Appendix~\ref{appendix:measure}. While we find surprisingly robust performance when varying $d$, we expect that there is no single best choice of diversity measure for all tasks.

The decision function $\Phi_{W,N,\alpha,d}$ can be interpreted as an unsupervised approach for producing labels of (near-)irreversibility. Taking this perspective, it is clear that these labels can be used alongside ground truth labels to supervise the learned irreversibility predictor from PAINT~\cite{xie2022ask}. Indeed the value of these labels is largely orthogonal to contributions from prior work and may be of value to work in RL safety and even in learning to ask for help~\cite{singh2022ask4help}.

\subsection{RM-RF with a Single Policy}\label{sec:single-policy}

Instead of the multi-policy forward-backward (FB) approaches frequently employed for RF-RL ~\cite{gupta2021reset,Achiam2017ConstrainedPolicyOpt,sharma2021earl,sharma2022state,xie2022ask} 
we propose to train single policy for RM-RL with randomly generated goals that are periodically changed during training. The intuition behind this is straightforward: to enable generalization we allow the agent to practice manipulating objects to many diverse goal positions, as these goals can be periodically changed during training without requiring a reset, this approach encourages the agent to experience explore the state space, improving generalization. 
This periodic goal switching is analogous to traditional, episodic, RL but with the key distinction that we we do not reset the agent's, or environment's, state when switching goals. Note that, perhaps surprisingly, not resetting the environment may actually result in the agent encountering a more diverse set of ``initial'' states (\ie states seen when given a new goal) than in the episodic setting as, in the episodic setting, the agent is frequently reset to some fixed initial position.  

We will now give a more formal comparison between our approach, FB-RL, and episodic methods. Specifically, recall that in traditional episodic RL, the objective to learn a policy that maximizes the discounted return \\[-1.5em]
\begin{align*}
    \pi^\star = \arg\max_\pi \mathbb{E}[J(\pi\mid g)] = \arg\max_\pi \mathbb{E}\left[\sum_{t=0} ^ \infty \gamma^t r(s_t, a_t\mid g)\right]
\end{align*}
~\\[-1em]
where, in comparison to the POMDP formulation from Sec.~\ref{sec:quantifying-irr}, we have, in the above, made the dependence on the randomly sampled goal $g\in\mathcal{G}$ where $\mathcal{G}$ is the set of all goals used during training.
In FB-RL, the ``forward'' goal space is normally defined as a singleton $\mathcal{G}_{f} = \{g^\star\}$ for the target task goal $g^\star$ (\eg the apple is on the plate, the peg is inserted into the hole, \etc). The goal space for ``backward'' phase is then the (generally limited) initial state space $\mathcal{G}_{b} = \cI \subset \cS$ such that $\mathcal{G}_f\cap \mathcal{G}_b=\emptyset$. As the goal spaces in FB-RL are disjoint and asymmetric, it is standard for separate forward/backward policies (with separate parameters) and even different learning objectives to be used when training FB-RL agents. In our setting, however, there is only a single goal space which, in principle, equals the entire state space excluding the states we detect as being NI (\ie, $\mathcal{G} = \mathcal{S} \setminus \{s_t\mid s_t\in\tau_{\pi} (t) \in \cT_\pi, \Phi_{W,N,\alpha,d} (\cT_\pi) = 1\}$). In our training setting, we call each period between goal switches a \textit{phase} and, when formulating our learning objectives, treat these phases as separate ``episodes'' in episodic approaches.

We provide extra details, pseudocode, and comparisons in Appendix~\ref{appendix:algo}.

\section{Experiments}\label{exps}
In our experiments, we look to answer a number of questions related to: (1) the importance of resets for learning, (2) how challenging \bench is for existing episodic and reset-minimizing approaches, (3) the efficacy of our proposed methodological contributions (unsupervised irreversibility detection and single-policy RM-RF) in reducing the number of resets required for learning and enabling out-of-distribution generalization, and (4) how our methods may be applied more generally to existing embodied tasks. 
To answer these questions, we run our experiments on three tasks in different embodied settings: our language-conditioned mobile manipulation \bench task, the Sawyer Peg task~\cite{yu2019meta,sharma2021autonomous}, and the RoboTHOR Object Navigation (ObjectNav) task~\cite{Deitke2020RoboTHORAO}. The Sawyer-Peg task, described in further detail below, is a popular stationary pick-and-place task used by prior work studying RF-RL and RM-RL. Our Sawyer-Peg experiments primarily use RGB observations rather than the low-dimensional observations used in most prior work. We include this environment to both better compare to prior work and to show the generality of our proposed contributions. Furthermore, to the best of our knowledge, we are the first to utilize methods for autonomous reinforcement learning for  ObjectNav, and demonstrate encouraging results. Before moving to our experimental results, we will first provide additional details of our evaluation environments, agent model architectures, and competing baselines.

\subsection{Environments}

\noindent \textbf{\bench}: See Section~\ref{sec:benchmark}.

\noindent \textbf{Sawyer-Peg}: In the Sawyer-Peg task, a Sawyer robot arm is attached to a table and must move a peg object sitting on a table into a hole within a box. As we are interested in the RM rather than RF learning, we remove the barriers from the table during training. So as to allow for evaluating generalization, during testing we consider a setting where \textit{both} the box and target hole are moved into novel positions (see Fig.~\ref{fig:benchmark}, center). This evaluation setting can be seen as being \textsc{Pos-OoD} with and (weakly) \textsc{Obj-OoD}. Visual observations include wrist-centric and third-person RGB images as in ~\cite{hsu2022vision,Sharma2023SelfImproving}. See Appendix~\ref{appendix:env:sawyer_peg} for more further details.

\noindent \textbf{ObjectNav}: To highlight the general applicability of our proposed RM-RL approach for embodied AI tasks, we show initial results on the RoboTHOR Object Navigation (ObjectNav) task~\cite{Deitke2020RoboTHORAO}. In ObjectNav, an agent (a LoCoBot robot) is placed within a simulated household environment and given a goal object category (\ie TV, chair, table, \etc), see Fig.~\ref{fig:robothor-objectnav}. The agent must then navigate to an object of this category using \textit{only} visual observations. The agent is considered to have successfully completed the task if it executes a special \textsc{End} action and an object of the given category is both visible and within 1m of the agent. During evaluation, must perform the same task, with the same set of object categories, but within unseen household environments (see Fig.~\ref{fig:benchmark}, right). This can be viewed as scene out-of-domain (\textsc{Scene-OoD}) evaluation. 
See Appendix.~\ref{sec:objectnav} for more details.

\subsection{Training Details}
All models are trained using the Proximal Policy Optimization~\cite{Schulman2017PPO} RL algorithm implemented within the AllenAct~\cite{Weihs2020Allenact} RL training framework. When training models in Sawyer-Peg (visual), we use 8 parallel processes and a rollout length of 300 for each, training for 1M steps requires approximately 20 minutes with checkpointing every 100k steps. When training models in \bench, we use 16 parallel processes with a rollout length of 200 for each, with 1M steps taking approximately 1.5 hours with checkpointing every 125k steps. Additional training details can be found in Appendix.~\ref{appendix:training_details}. 

\subsection{Model Architecture}

\begin{figure}[t]
    \centering
    \includegraphics[width=\linewidth]{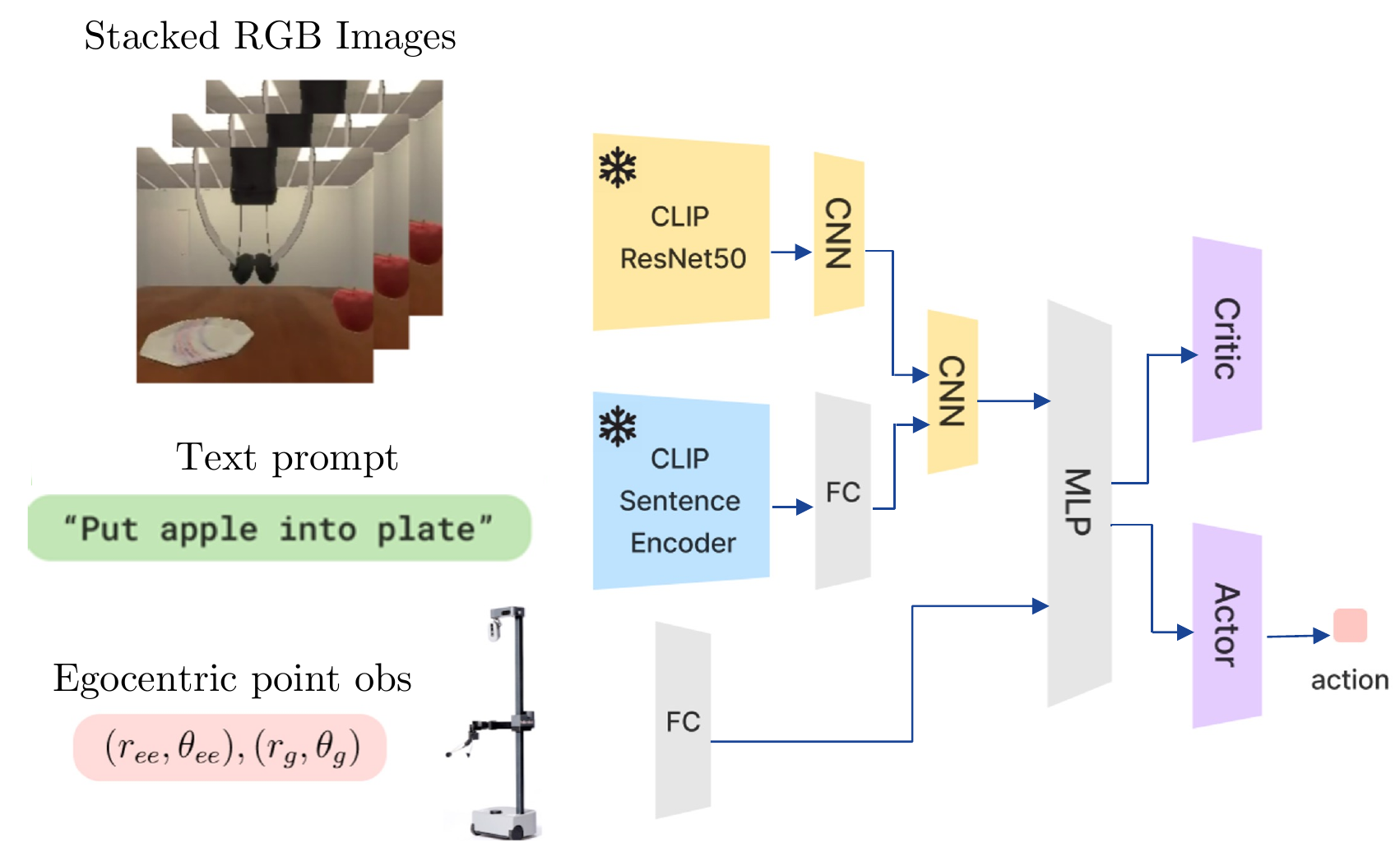}
    \caption{\textbf{\bench Model Arch.}}\label{fig:architecture}
    \vspace{-1.5em}
\end{figure}

As we train our models using the PPO algorithm, our agent architectures are of the actor-critic variety. For fair comparison across training strategies, we use the same model architectures across all baseline models (though this we use different architectures across tasks).
A diagrammatic representation of our model architecture used in \bench can be found in Fig.~\ref{fig:architecture}. Note that we use frozen CLIP~\cite{Radford2021CLIP} models to encourage visual generalization and language understanding. The model used for Sawyer Peg is similar to that for \bench but does not use CLIP. Unlike separately encoding two views of images in~\cite{hsu2022vision,Sharma2023SelfImproving}, we only use \textit{single} CNN visual encoder
that digests both views for parameter-efficiency. We find that single visual encoder is sufficient for solving this task. In ObjectNav, we use the same ResNet50 CLIP architecture with only egocentric visual observation input as proposed in~\cite{Khandelwal2022SimpleButEffective}. See Appendix~\ref{appendix:arch} for more details and hyperparameters.

\begin{figure*}[htbp!]
    \centering
    \includegraphics[width=1\linewidth]{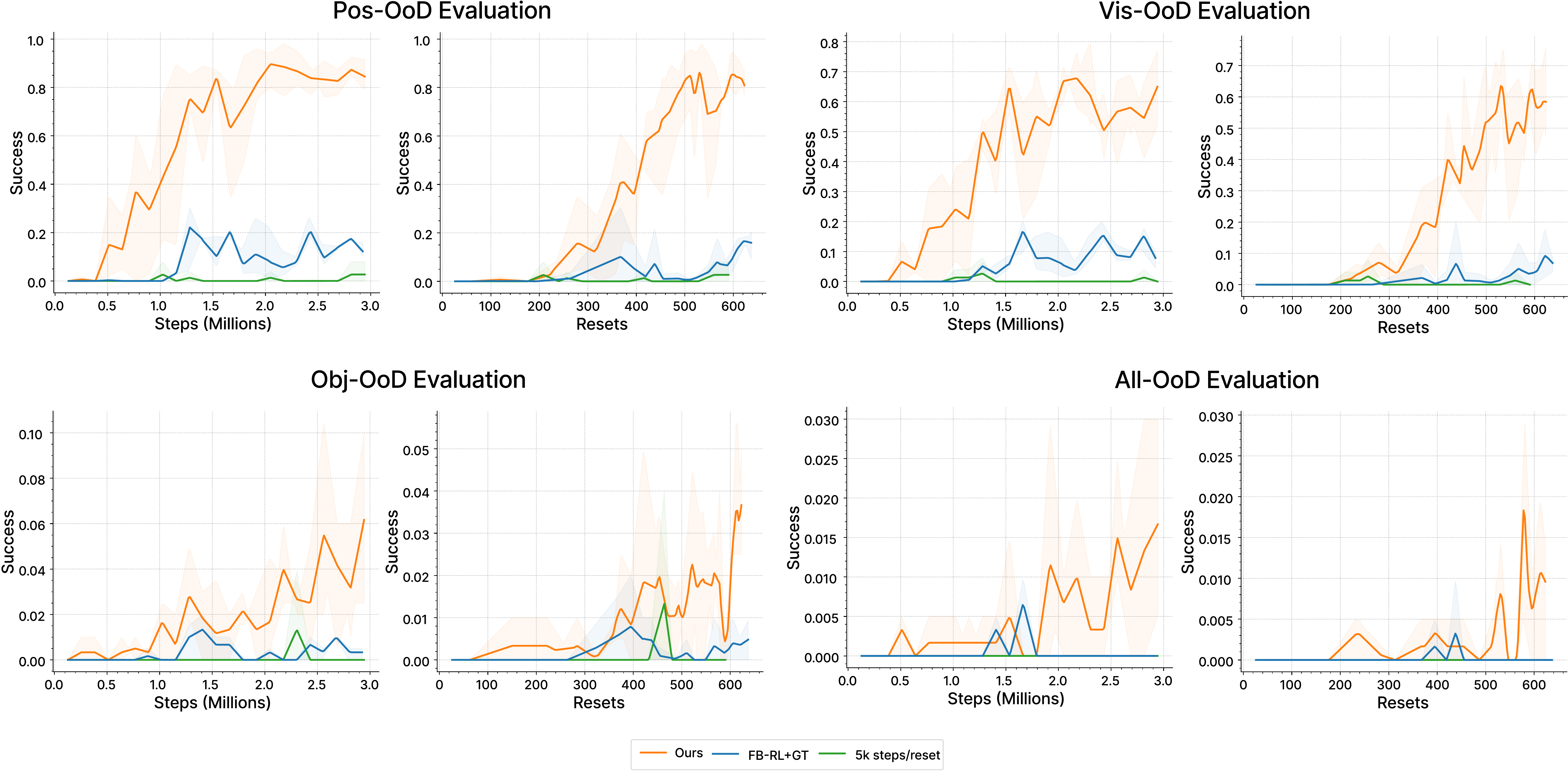}
    \caption{\textbf{\bench evaluation results for various methods across at different stages of training.}.}\label{fig:rm-pp-eval-results}
\end{figure*}

\newcolumntype{C}{>{\centering\arraybackslash}X}
\begin{table*}[htbp!]
\begin{minipage}{0.4\linewidth}
    \centering
    \includegraphics[width=0.8\linewidth]{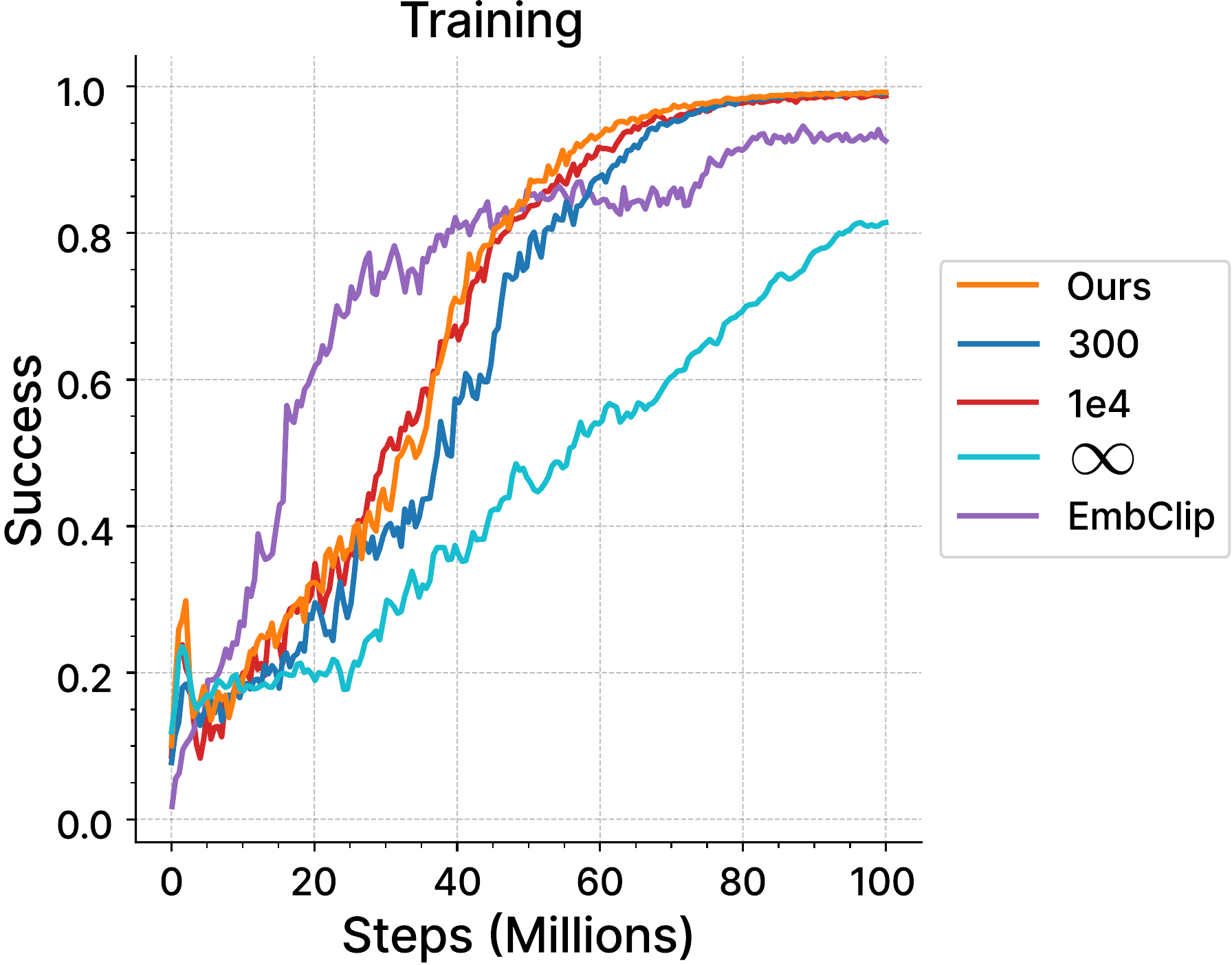}   %
    \captionof{figure}{\textbf{ObjectNav Training Curve}.}
    \label{fig:objnav_train}
\end{minipage}
\begin{minipage}{0.6\linewidth}
    \centering
    \small
    \begin{tabularx}{\linewidth}{C|CCC|CCC} \toprule
    \centering
       & Success ($50$M) & SPL ($50$M) & Resets ($50$M) & Success ($100$M) & SPL ($100$M) & Resets ($100$M)  \\ \midrule
    Ours        & 0.216 & 0.131 & \textbf{592 }               & \textbf{0.551}         &  \textbf{0.275}  &  \textbf{635} \\
    $N$=300       & 0.334       & 0.166         &  24k           &   0.355       & 0.167      &  1M \\
    $N$=10k       & 0.246       & 0.134    &  5k          &   0.418        & 0.218        &  10k \\
    $N$=$\infty$  & 0.206        & 0.141    &  \textbf{60}   &   0.339        & 0.178     &  \textbf{60} \\
        \cite{Khandelwal2022SimpleButEffective}  & \textbf{0.431 }      & \textbf{0.204}    &  1M            &   0.504       & 0.234        &  2M \\ \bottomrule
    \end{tabularx}
    \captionof{table}{\textbf{ObjectNav results for 50M and 100M steps.}}
    \label{tab:objnav_result}
\end{minipage}%
\end{table*}

\subsection{Baselines}
We consider the following three classes of baseline training strategies.

\noindent \textbf{Periodic resets ($+$ random goals).} Perhaps the simplest strategy for deciding when to request resets is simply to do so after every fixed number of steps.
Our periodic resets baselines do precisely this and are labeled simply as ``$N$ steps/reset'' where $N$ is some positive integer; here $N$ is set to generally be somewhat (or much) larger than in the standard episodic setting.
Note that, in principle, there are no irreversible states in ObjectNav as the task merely involves navigating around an, otherwise static, environment. For this reason, we also include a baseline trained without any resets beyond those used to initialize the environment, \ie $N=\infty$, and a baseline trained in the episodic setting just as in prior work~\cite{Khandelwal2022SimpleButEffective}. As we show below, choosing when to reset carefully can result in significant improvements in efficiency.

\noindent \textbf{FBRL+GT.} Here we implement the popular two-policy forward-backward training strategy from existing work. Inspired by PAINT~\cite{xie2022ask}, which learns a classifier trained on ground-truth irreversibility labels to request resets, we will use an \emph{oracle} version of this method and reset the environment whenever the agent enters one of a fixed collection of hand-labeled irreversible states (\eg, target object has fallen off the table). As was discussed in Sec.~\ref{sec:methods}, that this collection may not be exhaustive is a limitation of requiring ground truth labels.

\noindent \textbf{Ours (Random${+}$NI Measure).} We report results using our single-policy random-target training strategy with resets being requested based on our unsupervised irreversibility measures, recall Sec.~\ref{sec:quantifying-irr}. As different irreversibility measures may lead to different behavior and performance, we report multiple variants of our approach, one for each of our different irreversibility measures: \textsc{Std}, \textsc{Ent}, \textsc{DTW}, and \textsc{L2}. We describe the details of these irreversibility measures, which differ only based on the state diversity function used, in Appendix.~\ref{appendix:measure}.

Further details of baselines for each task can be found in Appendices~\ref{appendix:extended_bench_exps}, ~\ref{appendix:extended_sp_exps}, and~\ref{sec:objectnav}.

\subsection{Results}
We will now describe our experimental results in the context of the questions they are designed to address.

\begin{figure}[t]
    \centering
    \includegraphics[width=\linewidth]{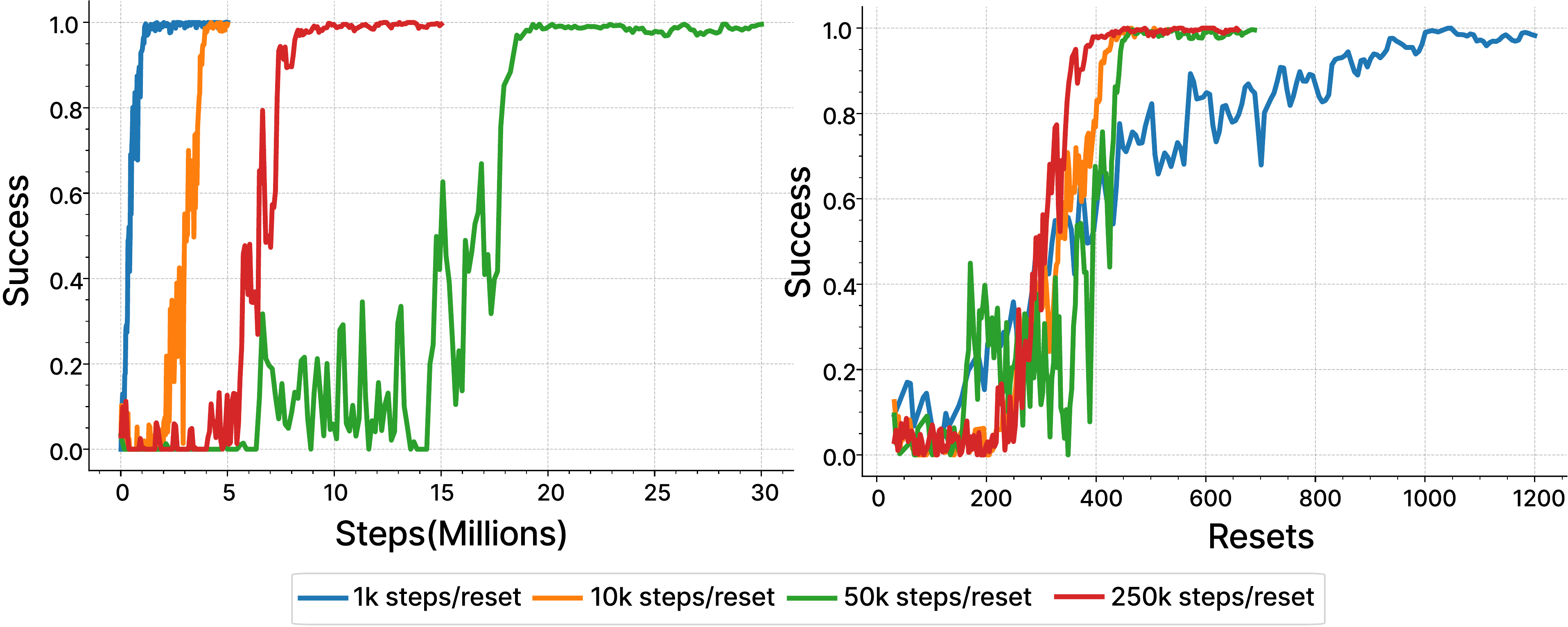}
    \vspace{-1em}
    \caption{\textbf{When controlling the number of resets, training curves converge.} Here we show the success of episodic baselines for the Sawyer-Peg task during training when plotting success against the number of training steps and, alternatively, against the total number of resets taken. In the left plot (Success \emph{v.s.} Steps) we see that methods with more frequent resets appear to substantially outperform, succeeding with far fewer total training steps. The right plot, however, tells a different story: when controlling for the number of resets, all methods learn at roughly the same rate. This suggests that resets are a fundamental part of what drives learning.
    }\label{fig:resets-important}
    \vspace{-1em}
\end{figure}

\noindent\textbf{How important are resets for learning?} In Fig.~\ref{fig:resets-important} we show training curves for various periodic reset baselines trained for the Sawyer-Peg task. In terms of training steps, the models which reset the most frequently are the most sample efficient with the models resetting less frequently taking millions of steps more to reach high performance. When plotting training performance against the number of resets requested, however, we find that the training curves begin to look very similar with surprisingly consistent trends despite noise inherent in RL training. This suggests that resets are of critical importance: in many cases, many training steps are effectively wasted, we hypothesize that the agent has undergone near-irreversible transitions, as the agent waits for a reset in the future that will enable it to begin learning again. We provide further evidence and point cloud visualizations for this hypothesis in Appendix~\ref{appendix:ni_sawyaer_peg}. Thus, as discussed in Sec.~\ref{sec:single-policy}, in the RM-RL setting it is important to evaluate agents considering both their sample and reset efficiency jointly. %

\begin{figure}[htbp!]
    \centering
    \includegraphics[width=1\linewidth]{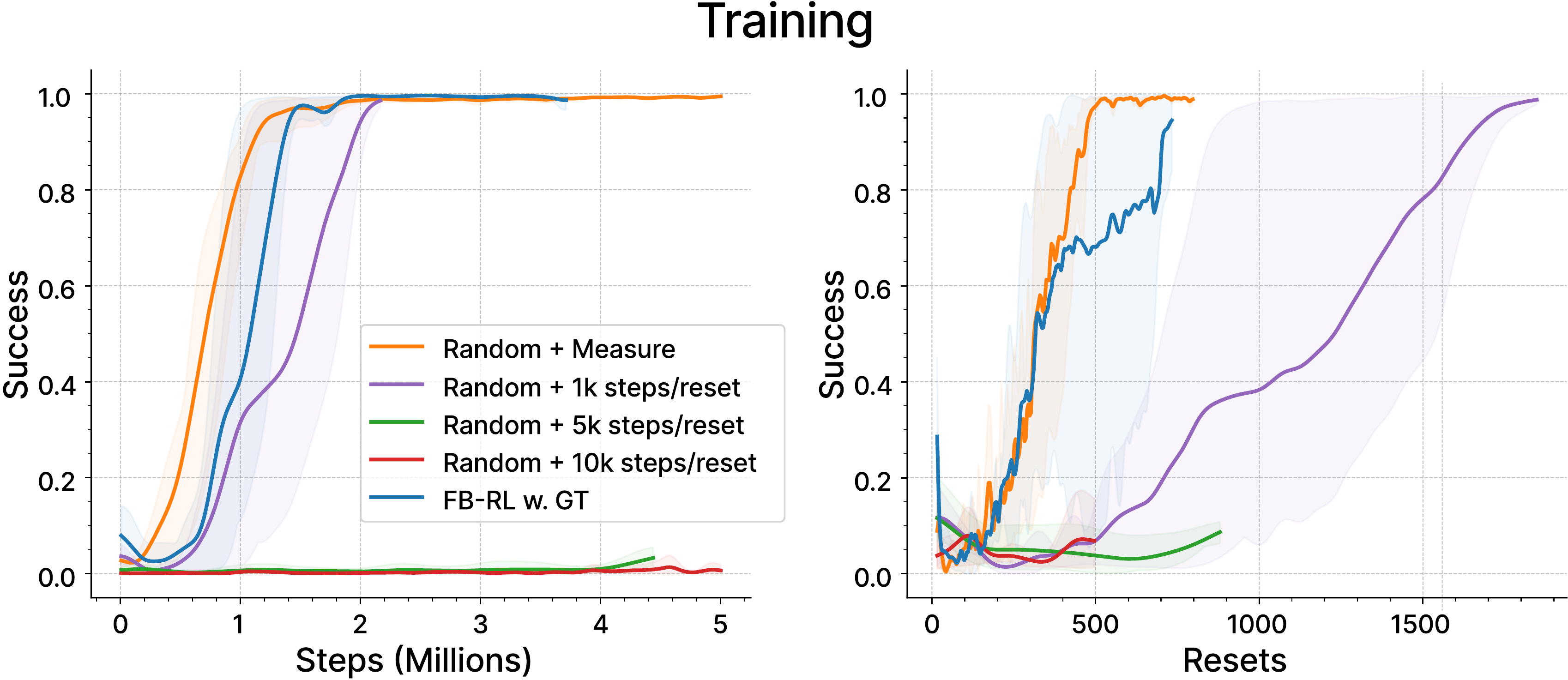}
    \caption{\textbf{\bench training performance.}}\label{fig:stretch_method_train}
    \vspace{-1em}
\end{figure}

\begin{figure}[htbp!]
    \centering
    \includegraphics[width=\linewidth]{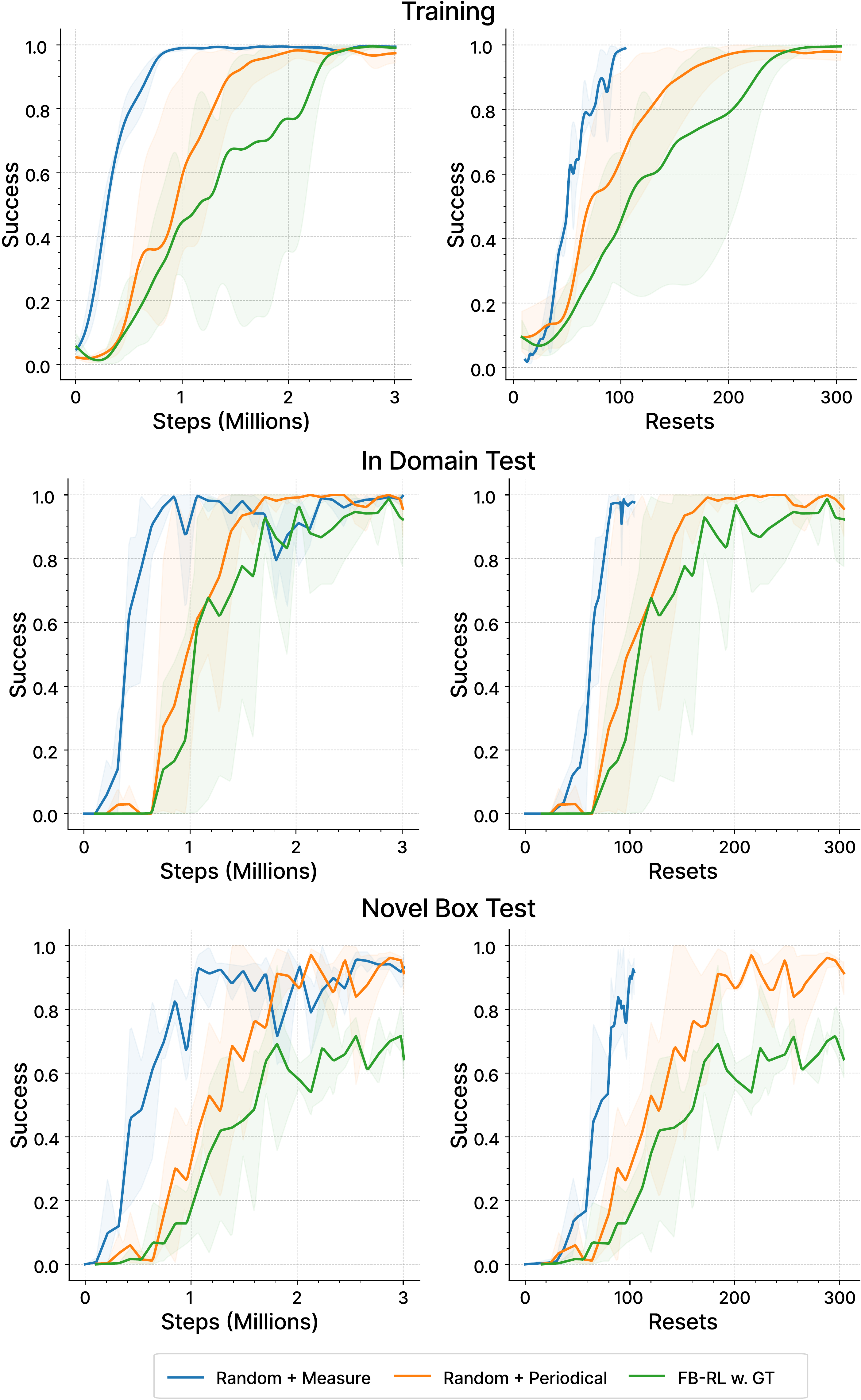}
    \caption{\textbf{Sawyer-Peg Results.}}
    \label{fig:sawyer_peg_method_comparison}
\end{figure}

\noindent\textbf{Can our unsupervised NI be used to reduce the number resets needed to train performant models?} Figure~\ref{fig:stretch_method_train} shows the training performance of our method versus other competing baselines for our \bench task with an object budget of 1. We find that our proposed approach is far more efficient in its use of resets: it achieves high success rates more consistently and with far fewer resets than the periodic reset baselines. Surprisingly our method is also \emph{more efficient in terms of training steps}. This suggests that our measures of NI transitions can consistently and accurately identify time-points where a reset will be of \textit{high value for learning}. Despite FB-RL having a set of possible goal states that is more constrained than our method (thus, intuitively, being easier to learn) we find that this constrained goal space appears to have little impact: even when controlling for resets, our method converges to high training performance as, or more quickly, than the FB-RL baseline (see Fig.~\ref{fig:stretch_method_train}). As we will discuss in more detail below, despite FB-RL having a similar (if somewhat slower) rate of training convergence, our method generalizes far more effectively. Similar trends hold for Sawyer Peg task, see the first row of Fig.~\ref{fig:sawyer_peg_method_comparison}.

\noindent\textbf{How well do RM-RL agents generalize?} 
We now provide evaluation results for testing the different facets of agent generalization proposed in Sec.~\ref{sec:benchmark} for \bench: \textsc{Pos-OoD}, \textsc{Vis-OoD}, \textsc{Obj-OoD}, and \textsc{All-OoD}. We also include results for novel box and hole positions for Sawyer Peg. We find, see Fig.~\ref{fig:rm-pp-eval-results}, that our methods handily outperform competing baselines across all evaluation settings. 
All evaluations are done for 200 randomly sampled tasks for models saved throughout training (for RoboTHOR ObjectNav we evaluate across the entire validation set which includes 1800 tasks from 15 unseen houses~\cite{Deitke2020RoboTHORAO}). Notably, for \textsc{Pos-OoD}, the easiest evaluation setting in \bench, our method experiences little to no drop in performance. We attribute this success largely to our random target strategy which results the agent experiencing a diverse set of potential goals. Competing baselines, \eg FB-RL$+$GT and agents using periodic resets, on the other hand show significantly worse performance on \textsc{Pos-Ood} than during training. This is despite the fact that, in \textsc{Pos-OoD}, the position of the physical receptacle remains the \textit{same for all baselines}. In Sawyer Peg, we observe similar trends, see Fig.~\ref{fig:sawyer_peg_method_comparison}, but with somewhat smaller drops in performance for the competing baselines; we attribute this to the smaller state space of Sawyer Peg which makes generalization somewhat easier. Surprisingly the generalization performance given cosmetic augmentations (\textsc{Vis-OoD}) is substantially better than the performance with novel object instances (\textsc{Obj-OoD}, \textsc{All-OoD}).  We suspect that this is, in part, due to our use of a frozen CLIP backbone encoder which is largely invariant to such cosmetic differences. Further discussion regarding these results can be found in Appendix.~\ref{appendix:exp:stretch_eval}.

\noindent\textbf{How does performance vary using different NI measures and object budgets?} 
For brevity and clarity, we provide ablation results for different choices for our proposed NI measures, and for varying object budgets (recall Sec.~\ref{sec:benchmark} and Fig.~\ref{fig:benchmark}) for \bench in Appendix.~\ref{sec:ablating-measures-stretch} and Fig.~\ref{fig:stretch_measure_ablation}, and for Sawyer Peg in Appendix~\ref{sec:ablating-measures-sp} and Fig~\ref{fig:sawyer_peg_measure_ablation}. In general, we find that our results are quite robust across diversity measures. We also find that the evaluation results for the \textsc{Obj-OoD} and \textsc{All-OoD} settings in \bench gradually increase when using larger object budgets but, perhaps surprisingly, larger budgets result in lower performance for \textsc{Pos-OoD} and \textsc{Vis-OoD}. We suspect that this may be because learning how to manipulate a single object is easier and allows for the agent to learn more complex manipulation behavior within the limited training time. See Appendix~\ref{appendix:extended_bench_exps} and~\ref{sec:ablating-measures-stretch} for further discussions.

\noindent\textbf{Is our RM-RL pipeline general and applicable to different embodied domains?}
We show our results of training RoboTHOR ObjectNav agents in Fig.~\ref{fig:objnav_train} with evaluation results in Table~\ref{tab:objnav_result}. The results are very promising: after 100M steps with only 635 resets we are able to achieve success rates \textbf{higher} than all competing baselines despite the next best performing baseline using 2M resets. Note also that our agent takes the vast majority (592 of 635) of its resets within the first 50M steps of training showing that our model continues to learn (going from a success rate of 0.216 at 50M training steps to 0.551 at 100M training steps) using very few resets. See Appendix.~\ref{sec:objectnav} for more details.

\section{Conclusion}
In this work we study the problem of training Reset-Minimizing Reinforcement Learning (RM-RL) agents within visually complex environments which can generalize to novel cosmetic and structural changes during evaluation. We design the \bench benchmark to study this problem and find that two methodological contributions, unsupervised irreversible transition detection and a single-policy random-goal training strategy, allow agents to learn with fewer resets and better generalize than competing baselines. 
In future work we look to further explore the implications of our irreversible transition detection methods for improving RM-RL methods and for building models that can ask for help during evaluation.
We also leave the space for design and balancing of how to penalize visits to unexpected NI states (with labels provided by our method), which may potentially conflict with encouraging exploration, as future work.

\section*{Acknowledgements}

We thank Winson Han, Eli VanderBilt, Alvaro Herrasti, and Kiana Ehsani for their assistance in the use of the AI2-THOR Stretch RE1 agent and in designing our experimental environment. We also thank Aniruddha Kembhavi and David Forsyth for helpful discussions during this project.

{\small
\bibliographystyle{ieee_fullname}
\bibliography{submission}
}

\clearpage
\appendix

\setcounter{figure}{0}
\renewcommand{\thefigure}{A.\arabic{figure}}
\setcounter{table}{0}
\renewcommand{\thetable}{\thesection.\arabic{table}}

\setcounter{figure}{1}
\renewcommand{\thefigure}{\thesection.\arabic{figure}}

\section*{Summary of Appendices}
\noindent These appendices include:
\begin{enumerate}
    \item[\ref{appendix:extended_exps}] Additional experimental results, analysis, and details.
    \item[\ref{appendix:env}] Low-level details of our experimental environments.
    \item[\ref{appendix:impl}] Implementation details for our irreversibility measures, model architectures, and training pipelines.
    \item[\ref{sec:objectnav}] Here we present preliminary but highly promising results when training Object Navigation agents in a reset-minimizing setting using our proposed approach.
\end{enumerate}

\section{Extended Experiments}\label{appendix:extended_exps}

\subsection{Additional \textbf{\bench} Details \& Analysis}\label{appendix:extended_bench_exps}

\subsubsection{Baseline details}

We first provide additional details regarding the differences between the baselines included in our experiments. ~\\[-0.8em]

\noindent$\bullet$ \textbf{Ours: random targets $+$ measurement-lead interventions}: Trained using random targets (see Sec.~\ref{appendix:algo} for algorithm details) and using either a (1) dispersion-based measure (\textsc{Std}, \textsc{Ent}), or a (2) Distance-based measure (\textsc{L2}, \textsc{Dtw}). See Sec.~\ref{appendix:measure} for the details.
~\\[-0.8em]

\noindent$\bullet$  \textbf{Random targets $+$ periodic interventions}: Trained using random targets with periodic resets taken every (1) $1$k, and (2) $5$k steps. No further resets are given. ~\\[-0.8em]

\noindent$\bullet$ \textbf{FB-RL $+$ GT}: FB-RL with both the periodic interventions and oracle explicit irreversible interventions (\eg the apple is dropped off the tabletop). This can be considered a ``ground-truth`` variant of PAINT~\cite{xie2022ask} as the intervention predicted by trained classifier is replaced with the oracle reset immediately after an object is dropped off the table. We also tried to use similarly scaled discrete penalties for visiting irreversible states illustrated in~\cite{xie2022ask} and found no significant difference.

We use an object budget of $1$, \ie the single task with the prompt ``\texttt{Put red apple into stripe plate}." for consistent comparison across baselines, and budget of $1, 2, 4$ for ablation of our method. Therefore, when using random targets during training the agent is told to manipulate the red apple to \textit{any} point goal \textit{over} the table, whereas FB-RL aims to move the apple back and forth from the plate to \textit{pre-defined} initial states \textit{on} the table. The targets, both randomly sampled and from pre-defined forward-backward states, are switched every $300$ steps during training. ~\\[-0.8em]

We now provide detailed discussions of our experimental results during training and evaluation.~\\

\subsubsection{\bench Training Performance}

We first discuss additional results regarding the training-time performance and efficiency of our various baselines.

\noindent$\bullet$ \textbf{Reset and sample efficiency}: as shown in Fig.~\ref{fig:stretch_method_train}, our method achieves both high sample, and reset, training efficiency when compared to other baseline models.~\\[-0.8em]

\noindent$\bullet$ \textbf{Toward resetting only when necessary}: We provide an additional ablation when using more, or less, frequent periodic resets during training. As shown in Fig.~\ref{fig:stretch_method_train} simply increasing or decreasing the frequency of periodic resets does little to bridge the gap between our approach and these baselines. Resetting every 1k steps results in similar efficiency in terms of the training steps but requires 3${\times}$ times more resets. Alternatively, resetting every 5k steps results in similar a similar number of total resets after 5M training steps as our method but, unlike our method, results in a very poor success rate.
~\\[-0.8em]

\noindent$\bullet$ \textbf{Random targets introduce little additional difficulty over FB-RL}: Intuitively the forward-backward gameplay of FB-RL models should be easier to learn than when using random targets as the space of goal states of FB-RL is a small subset of those used when randomizing targets. However, even when we attempt to control for resets (similar reset rate for convergence in Fig.~\ref{fig:stretch_method_train}), this appears not to be the case. The runs trained with FB-RL, even with the immediate ground truth interventions for explicit irreversible states, exhibit slightly
slower training convergence than our method (see Fig.~\ref{fig:stretch_method_train}). Therefore, constraining the goal space during training appears to have little impact on simplifying the autonomous learning process. Moreover as, during deployment, we only care about the forward policy trained by FB-RL, it seems intuitively wasteful to spend half of the training time learning to autonomously reset instead of practicing more relevant tasks. Using a single policy with random targets makes every phase equivalent and thereby may result in more efficient use of training time.~\\

\subsubsection{\bench Evaluation Results}\label{appendix:exp:stretch_eval}

\begin{figure*}[t]
    \centering
    \includegraphics[width=1\linewidth]{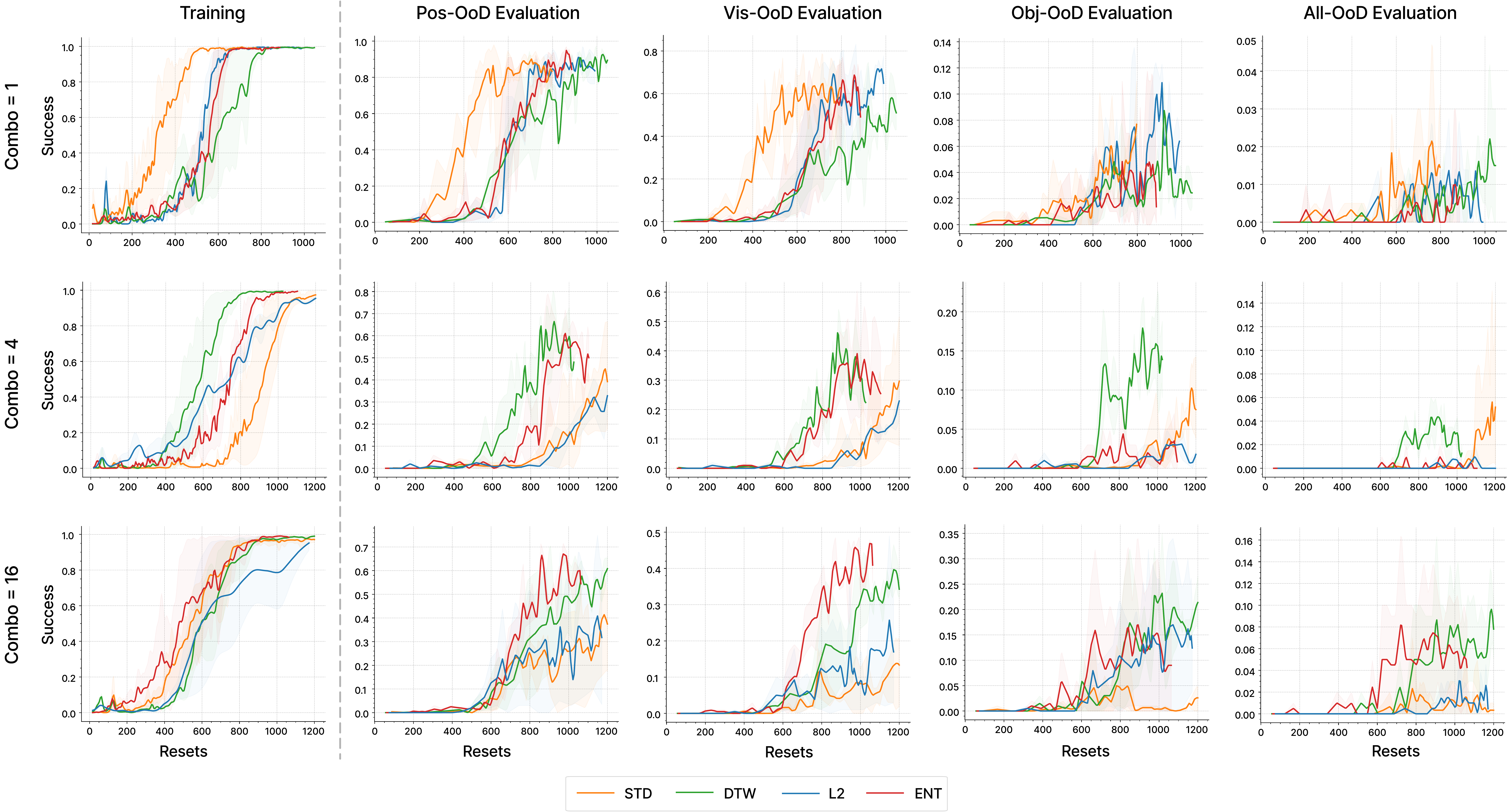}
    \caption{\textbf{\bench Irreversibility measures \& object budget ablations}. Our measurement-determined irreversibility reset approach is relatively robust to the selection of diversity measure. Note that all models are trained for 5M environment steps but the total number of resets taken by each method may differ.}\label{fig:stretch_measure_ablation}
    \vspace{-1em}
\end{figure*}

We now discuss additional evaluation-time results of our trained agents. In particular, we provide additional insight into the generalization abilities of our trained models. Recall, Sec.~\ref{sec:benchmark}, that we have four evaluation settings testing different facets of agent generalization: \textsc{Pos-OoD}, \textsc{Vis-OoD}, \textsc{Obj-OoD}, and \textsc{All-OoD}.

\noindent $\bullet$ \textbf{\textsc{Pos-OoD}}: In \textsc{Pos-OoD}, the picking household objects and target receptacles are placed randomly within reach on the table. Due to limited visibility, one or both objects may not be visible to the agent initially. Additionally, the Stretch robot's unique characteristics, which allows for vertical and horizontal movement but no change in arm pitch, make it difficult for the agent to differentiate between larger objects further away from the gripper and smaller objects closer to it. Our method achieves the highest degree of positional-invariance among all the methods, see Fig.~\ref{fig:rm-pp-eval-results}~(top left). We suspect that next best baseline, FB-RL${+}$GT, method was not able to generalize well due to the constrained space of goal states it observes during training. 
Interestingly, using an object budget 2 or 4, see Fig.~\ref{fig:stretch_measure_ablation}, results in lower performance compared to when using a budget of 1. We suspect that this may be because learning how to manipulate a single object is easier and allows for the agent to learn more complex manipulation behavior within the limited training time. ~\\[-0.8em]

\noindent $\bullet$ \textbf{\textsc{Vis-OoD}}: We found, see Fig.~\ref{fig:rm-pp-eval-results}~(top right), that our method can generalize to the unseen cosmetic changes, \ie novel textures (\eg texture and materials of the table, floor, wall, ceiling) and lighting (random light color and intensity). See Sec.~\ref{appendix:env} for randomization details.
We attribute this generalization ability in part to the pretrained CLIP backbone used in our model (see Appendix~\ref{appendix:arch} for architecture details). It is interesting that the performance drops while the object budget increases (see Fig.~\ref{fig:stretch_measure_ablation}); we suspect this is due the increase in budgets making training more challenging.

\noindent $\bullet$ \textbf{\textsc{Obj-OoD}}: All baselines performed relatively poorly in this setting, Fig.~\ref{fig:rm-pp-eval-results}~(bottom left), though the models with the largest object budget (a budget of 4) during training perform the best, see Fig.~\ref{fig:stretch_measure_ablation}. This is not surprising as the language backbone is completely frozen and the visual inputs are only raw RGBs. We suspect that it will be possible to achieve higher success if a perception module with more input streams or augmentations is used. For example, we might include depth images, object center positions estimated in pixel space (as in~\cite{zhang2023cherry}), detected object bounding boxes (as in~\cite{jiang2022vima}), or generative augmentations (as in~\cite{yu2023scaling}). As a baseline model, we kept the input as the raw RGB image without augmentation and we leave potential improvements as future work.

\noindent $\bullet$ \textbf{\textsc{All-OOD}}: For our most challenging evaluation setting the room structure (\eg furniture at the background), texture and lightning in \textsc{Vis-OoD}, and objects in \textsc{Obj-OoD}, are all unseen. Our results, see Figures~\ref{fig:rm-pp-eval-results}~(bottom right) and~\ref{fig:stretch_measure_ablation}, show that this setting is indeed more challenging than either \textsc{Vis-OOD} or \textsc{Obj-OOD} alone. As all methods struggle in the challenging setting there is still significant room for novel approaches to be designed to solve this task.

\subsubsection{Ablating Measures of Irreversibility on \bench}\label{sec:ablating-measures-stretch}

\begin{figure}[htbp!]
    \centering
    \includegraphics[width=0.8\linewidth]{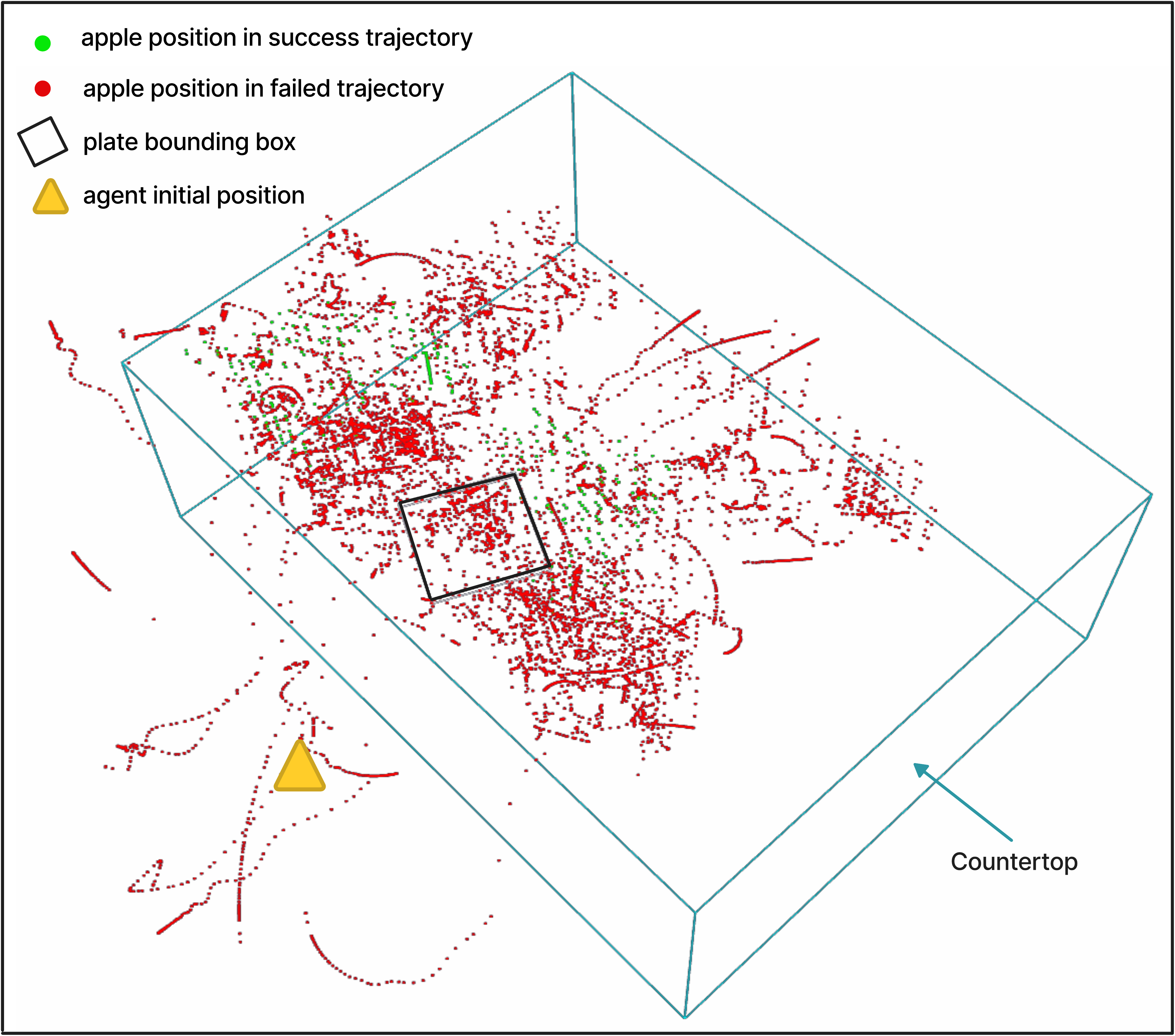}
    \caption{\textbf{Visualizing successful and failed object trajectories in \bench during training.} Notice that the object occupies many diverse states and can fall off of the table or roll away from the agent.}
    \label{fig:stretch_intermidate_point_cloud}
    \vspace{-1.5em}
\end{figure}

We now show that our measurement-determined irreversibility intervention method is relatively robust to the selection of diversity measure (recall the discussion in Sec.~\ref{sec:quantifying-irr}). To illustrate this, we used two classes of measures: (1) dispersion-based (\textsc{Ent}, \textsc{Std}) and (2) distance-based (\textsc{L2}, \textsc{Dtw}). We provide details and formulations for our proposed measurements in Appendix~\ref{appendix:measure}. As shown in Fig.~\ref{fig:stretch_measure_ablation}, no single measure outperforms the others across all evaluation settings. Note that objects occupy a diverse range of states during training, especially when the apple is dropped from the hand in the air. We visualize 10k such steps sampled through training in 3D space in Fig.~\ref{fig:stretch_intermidate_point_cloud}. 
The visualization shows that near-irreversible states, and not just explicitly irreversible states (\eg the apple falling from the table), bottleneck the autonomous training process. As our experiments show, our proposed irreversibility measures are able to characterize the near-irreversible states and allow for interventions, namely resets, to be taken effectively.

\subsection{Additional Sawyer-Peg Details and Analysis}\label{appendix:extended_sp_exps}

We now provide additional results and analysis for our experiments within the Sawyer-Peg environment.

\subsubsection{Experiments with RGB Observations} \label{sec:sawyer-peg-rgb}

Unlike previous work with only experiments on low-dimensional observations, we extend the task into the visual domain and found our method could easily solve the problem and make good generalizations. We show our experiment results at Fig.~\ref{fig:sawyer_peg_method_comparison}. We compare our method (which uses random targets and measurement-determined resets) with two categories of baselines: (1) random targets with the periodic resets (1e4), and (2) FB-RL with the periodic resets as well as the ground-truth resets if the peg is dropped off the table. Similarly as the results for \bench, our method achieves significantly more efficient training in terms of training steps and total resets. When evaluating on novel box goals, \ie where the target hole of the peg box has been randomized relative to the box (see Fig.~\ref{fig:benchmark} and Sec.~\ref{appendix:env:sawyer_peg} for details), we find that the FB-RL agents cannot generalize as well as agents trained with random targets. Therefore, the our proposed approach both makes training more efficient (by providing resets when necessary) and allows for more significant generalization (through the use of random targets during training).

\subsubsection{Ablating Measures of Irreversibility on Sawyer-Peg}\label{sec:ablating-measures-sp}

\begin{figure*}[htbp!]
    \centering
    \includegraphics[width=0.67\linewidth]{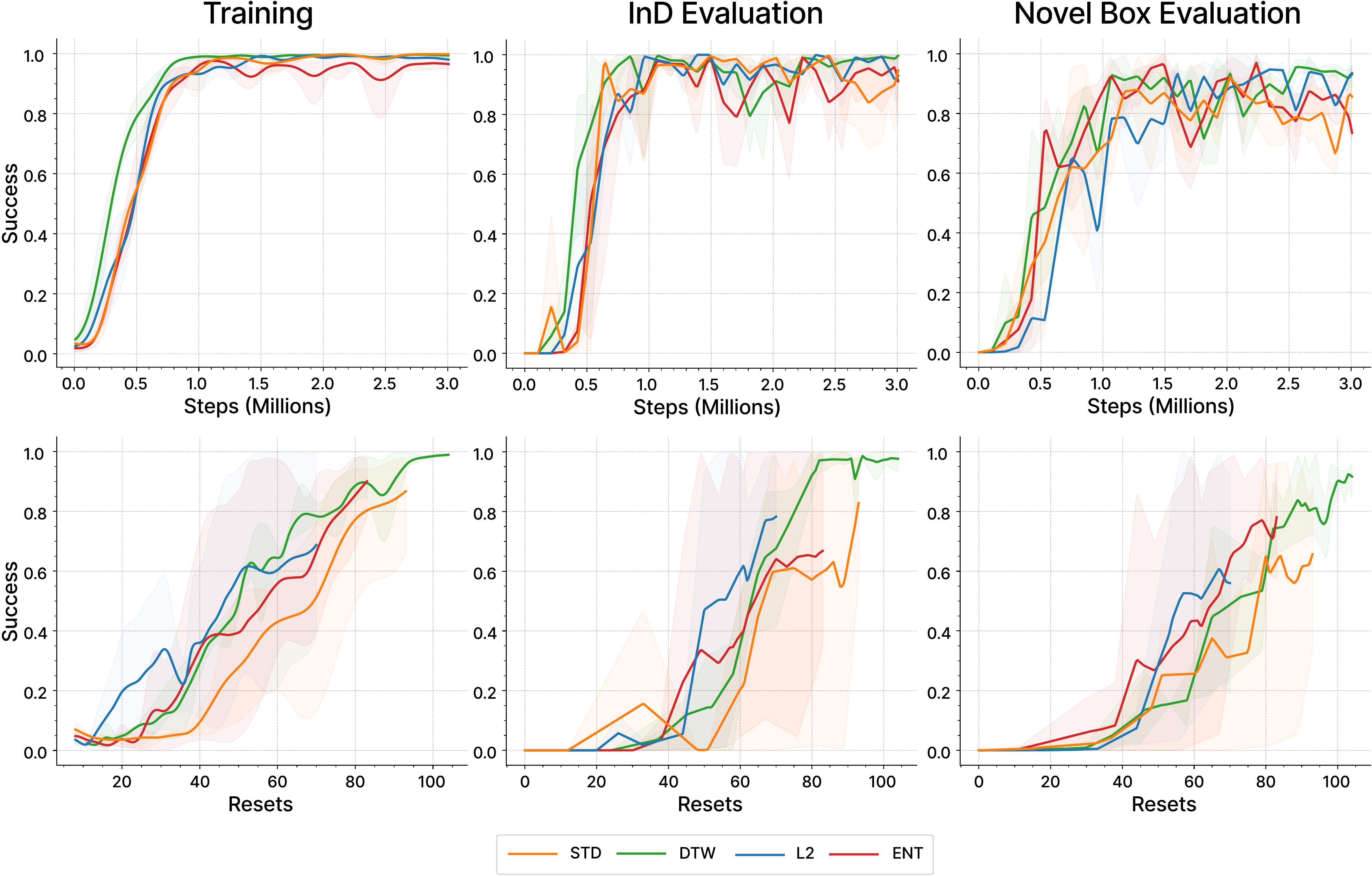}
    \caption{\textbf{Irreversibility measure ablation within the Sawyer-Peg environment.}}
    \label{fig:sawyer_peg_measure_ablation}
\end{figure*}

\begin{figure}[htbp!]
    \centering
    \includegraphics[width=\linewidth]{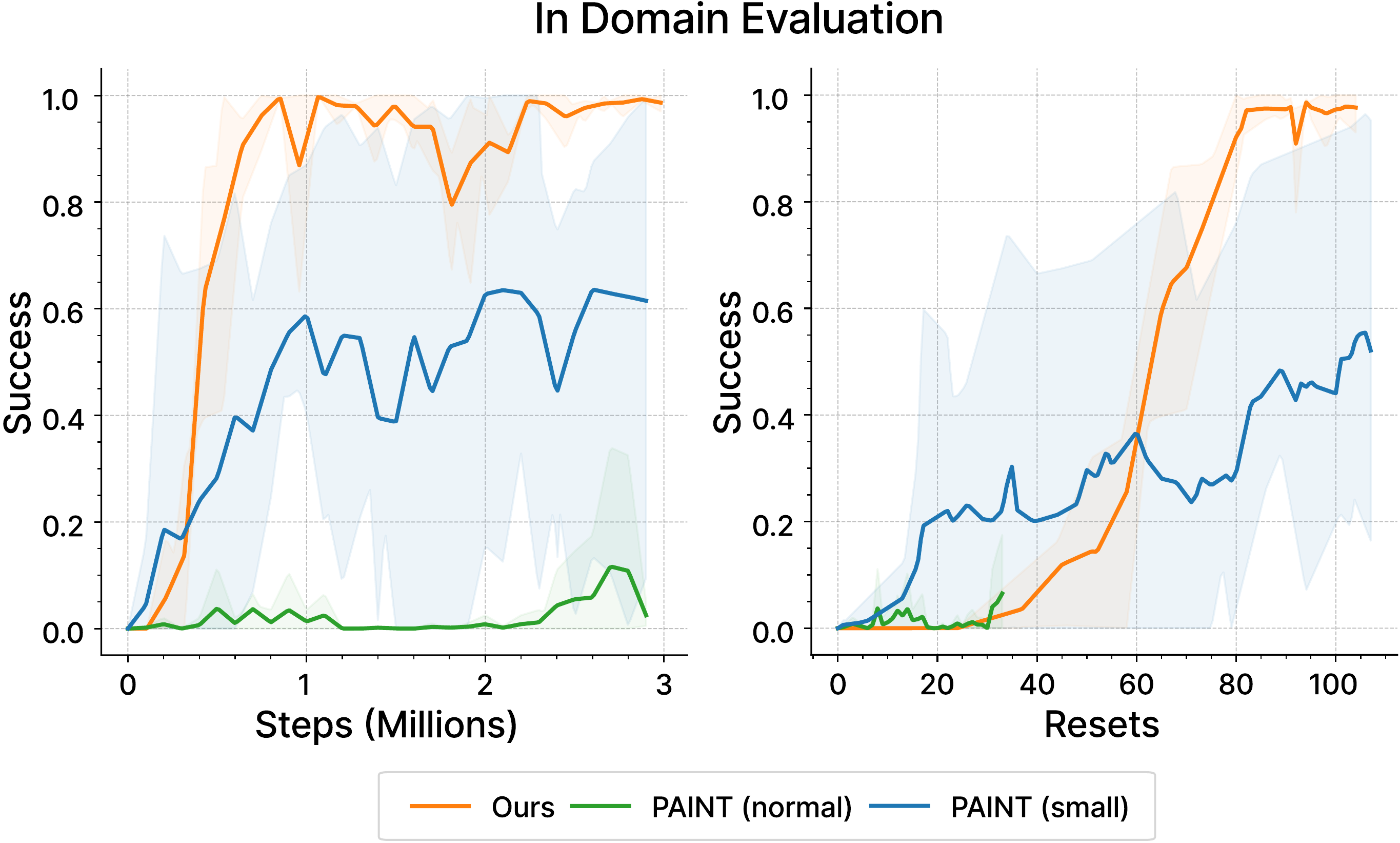}
    \vspace{-2em}
    \caption{\textbf{Evaluation with supervised method PAINT~\cite{xie2022ask} trained in narrower table and the normal size table (low-dim)}. The significant performance decrease shows for training with the normal size table, mainly due to the false negative labels of near-irreversible states used for training classifiers in PAINT. Due to high oscillations, PAINT experiments are trained with five random seeds and all methods are evaluated with 200 tasks and shadowed area represent the min/max range}
    \label{fig:sawyer_peg_paint_comparison}
\end{figure}

Similarly as in Section~\ref{sec:ablating-measures-stretch}, we provide additional ablations in the Sawyer-Peg environment showing the relative performance of baselines trained using different irreversibility measures in Fig.~\ref{fig:sawyer_peg_measure_ablation}. Note that \textbf{all} of our proposed baselines achieve consistently high performance in the training, in-domain evaluation, and novel box evaluation. Moreover, they achieve this high performance within $\bm{{\approx}100}$ \textbf{resets} in total and converge in around \textbf{1M steps}. All experiments are run with three random seeds and shaded areas represent the min/max range across different seeds. Note that previous benchmark EARL~\cite{sharma2021earl} shows almost 0 success in a variant of this task using FB-RL with periodic resets over 3 to 5 million steps even when: (1) the evaluation environment is identical to that during training, (2) low-dimensional observations containing the true position of the peg are given to the agent, and (3) table boundaries are present to prevent the peg from dropping from the table. More recent work introduced from~\cite{xie2022ask} use a much narrower, boundaryless, table for this task and achieve approximately 60\% success in 3M steps with 120 hard resets on average. We reproduce the result at Fig.~\ref{fig:sawyer_peg_paint_comparison}.

\subsubsection{Existence of Near-Irreversible States in Sawyer-Peg}\label{appendix:ni_sawyaer_peg}

Here we provide further evidence suggesting that prior, reset-free, works' relatively low performance in the Sawyer-Peg task is largely due to the existence of near-irreversible states.

\begin{figure}[h]
    \centering
    \includegraphics[width=\linewidth]{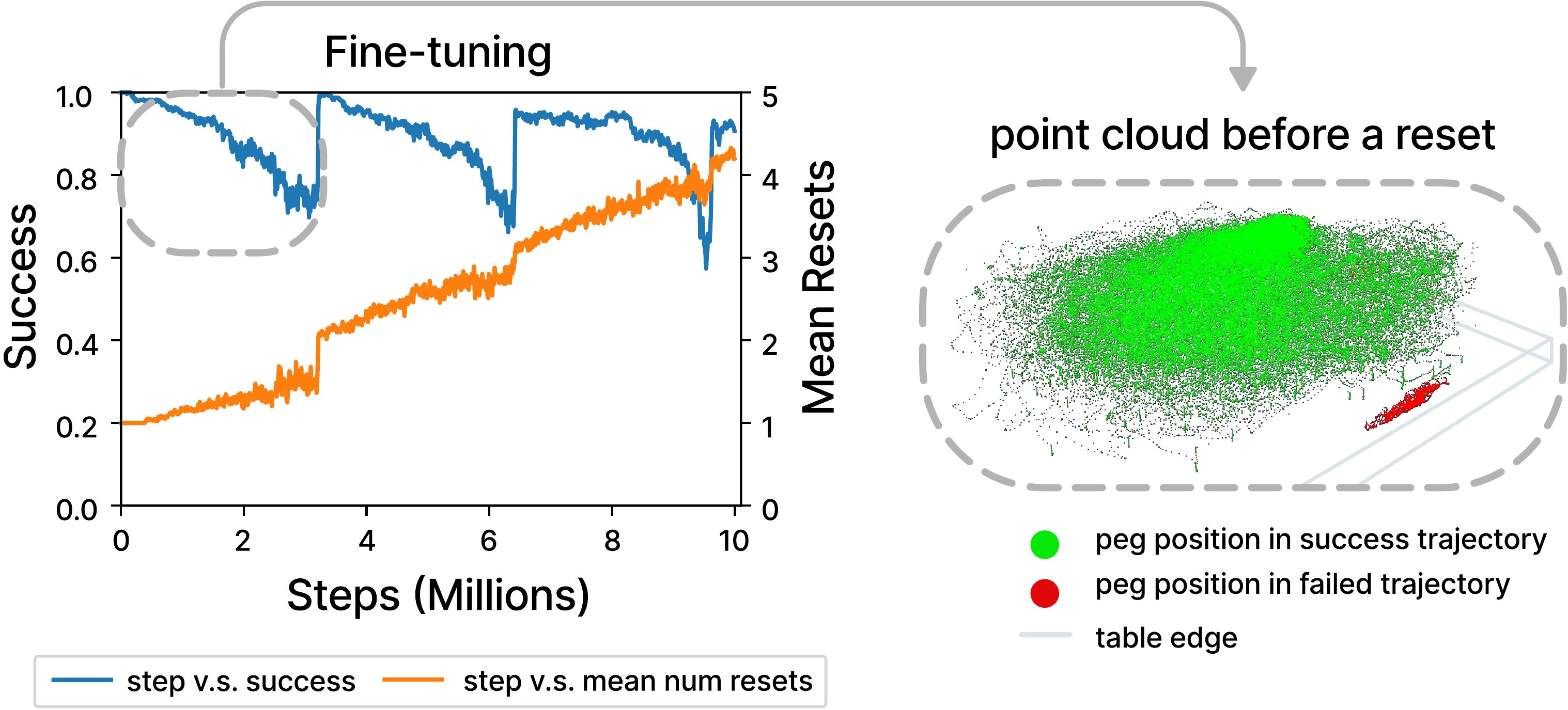}
    \caption{\textbf{Finetuning a Sawyer-Peg agent in a low-reset setting.} A well-trained Sawyer-Peg agent is finetuned in the same environment but with periodic resets happening much less frequently than in the episodic setting (approximately every 3.5M steps). Note that, beyond periodic resets, we reset the environment whenever the agent drops the peg off of the table so the mean number of resets per process will steadily increase during training. Agent performance appears to steadily decrease until a periodic reset occurs after which it returns to a success rate near 100\%. Note that periodic resets occur across all processes simultaneously.}
    \label{fig:sawyer_peg_ft_failure}
\end{figure}

\noindent \textbf{RF Fine-tuning Failure} 
To this end, we train an episodic agent in the random targets setting and then fine-tune this agent \textbf{in the same setting and environment} but with using much less frequent resets and a small learning rate. Note that, by the end of the initial stage of training (\ie before finetuning) the agent achieves a near 100\% success rate. Beyond periodic resets, during finetuning we provide additional resets immediately after the peg dropped off the table. We provide these resets so we do not conflate the impact of ``true`` irreversible states with near-irreversible states. As shown in Fig~\ref{fig:sawyer_peg_ft_failure}, where blue curve shows the success rate, and the orange curve shows the number of resets \textit{on average} for each process during fine-tuning, the agent achieves a near $100\%$ success rate at the start of fine-tuning. However, this success rate \emph{continuously decreases} until a periodic resets occurs after $4M$ training steps after which the success rate suddenly recovers back to near $100\%$. Note that we have already provided ``oracle'' resets after peg dropped to exclude the ``true'' irreversible states and yet, the agent's success rate still decreases. This suggests that the agent manages to enter environment states where the peg is still on the table but from which obtaining success is difficult, \ie the agent enters near-irreversible states. To provide a qualitative visualization of where the agent fails during finetuning, we track all positions of the peg in one environment before the first reset and label them red and green for failed and successful trajectories respectively (see Fig~\ref{fig:sawyer_peg_ft_failure}). As showing in the 3D point cloud on the right, the success and failure has a clear dividing line near the edge of the table. Thanks to the random targets training setting, the agent can succeed from a diverse range of states but those states near the edge of the table appear to be very difficult (perhaps due to physical limitations on the degrees of freedom of the arm) and thereby represent a set of near-irreversible states.

\begin{figure*}[h]
    \centering
    \includegraphics[width=\linewidth]{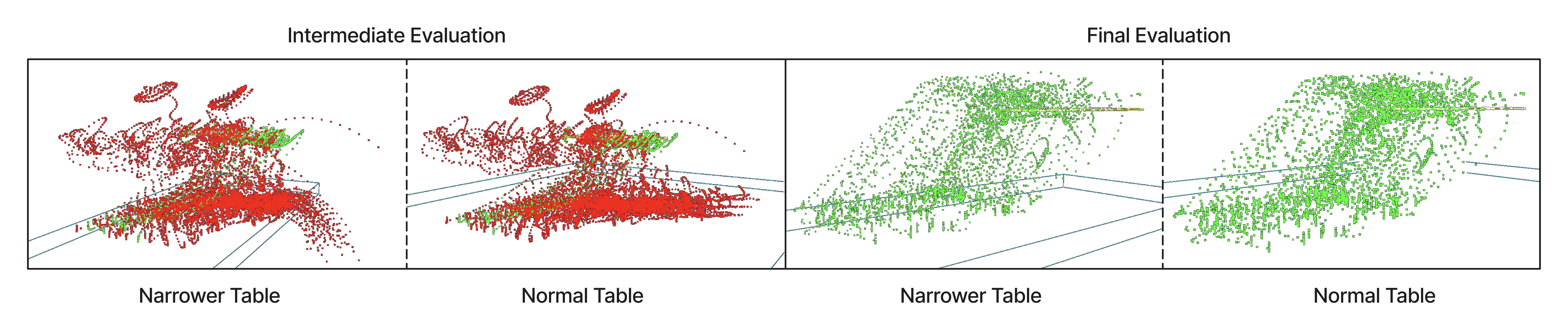}
    \vspace{-2em}
    \caption{Point cloud visualizations for evaluations on the narrower (same as~\cite{xie2022ask}) and the normal-sized table, where red points indicate the object (peg) head positions for failed trajectories while greens show the successful ones. Evaluations on two settings have the exact the same performance and rollouts trajectories for the final checkpoint (last two figures), but the same policy is leading to different consequences during the evaluation at 300k steps (first two figures): the peg always drops off the narrower table but mostly is still on the edge of the normal-size table.}
    \label{fig:sawyer_peg_point_cloud_small_table}
\end{figure*}

\noindent \textbf{Evaluation with Narrower Table}
    Besides, we provide further evaluations on the narrower table use in~\cite{xie2022ask}. We surprisingly found that the result is exactly the same as evaluating with the normal-sized table. We illustrate the point clouds of successful and failed trajectories evaluated intermediately and finally in Fig.~\ref{fig:sawyer_peg_point_cloud_small_table}, where red points indicate the object (peg) head positions for failed trajectories while greens show the successful ones. Evaluations on two settings have the exact the same performance and rollouts trajectories for the final checkpoint (last two figures), but the same policy is leading to different consequences during the evaluation at 300k steps (first two figures): the peg always drops off the narrower table but mostly is still on the edge of the normal-size table. We consider the second case as near-irreversible where we are not able to expect the agent to grasp the peg back given the current imperfect policy.

The above results motivate a need for the measures of irreversibility we introduce in this work: it is a \textit{practical impossibility} to label all (near-) irreversible states before training.

\section{Environment Details.}\label{appendix:env}
Besides the introduction in Sec.~\ref{sec:benchmark} and Appendix.~\ref{appendix:extended_bench_exps}, we further provide low-level details. Code made available at \href{https://zcczhang.github.io/rmrl}{\texttt{https://zcczhang.github.io/rmrl}}.

\subsection{\bench} \label{appendix:env:stretch}

\noindent \textbf{Observations.}
In experiments of \bench, the observation space consists of the visual observation from the wrist-centric camera with dimension $224\times224$ and field of view (FOV) of 75, the text prompts, and the low-dimensional observation including the egocentric coordinates of the Stretch gripper and the target (4-dimensional in total). The example is show in Fig.~\ref{appendix:arch}. The language instructions are in the format of "\texttt{Put \{obj\_1\} into \{obj\_2\}.}" for object goal and "\texttt{Put \{obj\_1\} into ($r, \theta$).}" for point goal in the polar form $r, \theta$. In \bench, the target task can be further decomposed as the picking task where the prompt is only "\texttt{Pick \{obj\_1\}.}" if there is no object in agent's hand.

\noindent \textbf{Action Space.}
The action space is discrete and consists of $10$ actions. The robot base has forward and backward movement in parallel to a side of the table along $x$ axis relatively (see Fig.~\ref{fig:irreversibility-types}). The robot arm can lift or lower along the relative $y$ axis (in AI2-Thor, the $y$-axis corresponds to the height, which is usually the vertical $z$-axis in Cartesian coordinate space); the arm can also extend or retract horizontally to move the gripper further or closer relative to the agent base in the $z$-axis; and the gripper can rotate in positive and negative yaw directions. For object interactions, we have a \texttt{PickUp} action where only the specified object with a unique object ID and within a sphere with radius $r$ centered at the gripper (similar to ManipulaTHOR~\cite{Ehsani2021Manipulathor}) will be grasped with its previous position and orientation unchanged in hand. The release action simply drops the object if there is one in hand, and we add simulation steps to wait until the object stabilizes for a realistic setting. Table~\ref{tab:stretch_action} provides details about the action names, descriptions, and scales used in \bench. For comparison, we also provide the action space for the RoboTHOR object navigation task in Table~\ref{tab:objnav_action}.

\begin{table*}[htbp!]
\centering
\small
\begin{tabularx}{0.8\linewidth}{@{}>{\raggedright\arraybackslash}m{0.18\linewidth}|>{\raggedright\arraybackslash}m{0.5\linewidth}|>{\raggedright\arraybackslash}m{0.15\linewidth}@{}}
\textbf{Action} & \textbf{Description} & \textbf{Scale} \\ \midrule
\texttt{MoveAhead} & Move robot base in $+x$ axis & 5 cm \\
\texttt{MoveBack} & Move robot base in $-x$ axis & 5 cm \\
\texttt{MoveArmHeightP} & Increase the arm height ($+y$) & 5 cm \\
\texttt{MoveArmHeightM} & Decrease the arm height ($-y$) & 5 cm \\
\texttt{MoveArmP} & Extend the arm horizontally ($+z$) & 5 cm \\
\texttt{MoveArmM} & Retract the arm horizontally ($-z$) & 5 cm \\
\texttt{MoveWristP} & Rotate the gripper in $+x$ yaw direction & 2$^\circ$ \\
\texttt{MoveWristM} & Rotate the gripper in $-x$ yaw direction & 2$^\circ$ \\
\texttt{PickUp(object\_id)} & Pick up object with specified unique \texttt{object\_id} if object within the sphere with radius $r$ centered at gripper & $r=0.06$ \\
\texttt{Release} & Release object with simulation steps until object is relatively stable & -- \\
\end{tabularx}
\caption{\textbf{Action space in \bench.} All directions are relative to the Stretch robot.}
\label{tab:stretch_action}
\end{table*}

\begin{table*}[htbp!]
\centering
\small
\begin{tabularx}{0.8\linewidth}{@{}>{\raggedright\arraybackslash}p{0.18\linewidth}|>{\raggedright\arraybackslash}p{0.5\linewidth}|>{\raggedright\arraybackslash}p{0.15\linewidth}@{}}
\textbf{Action} & \textbf{Description} & \textbf{Scale} \\ \midrule
\texttt{MoveAhead} & Move robot base in $+x$ axis & 0.25 m \\
\texttt{MoveBack} & Move robot base in $-x$ axis & 0.25 m  \\
\texttt{RotateLeft} & Rotate the robot base leftward (yaw) & 30$^\circ$ \\
\texttt{RotateRight} & Rotate the robot base rightward (yaw) & 30$^\circ$ \\
\texttt{LookUp} & Rotate the camera upward (pitch) & 30$^\circ$ \\
\texttt{LookDown} & Rotate the camera downward (pitch) & 30$^\circ$ \\
\texttt{End} & Terminate the episode. Success if the goal object within $d$ visibility. & $d=1$ \\
\end{tabularx}
\caption{\textbf{Action space in RoboThor ObjectNav.} All directions are relative to the LoCoBot.}
\label{tab:objnav_action}
\end{table*}

\noindent \textbf{Success Criteria.}
In \bench tasks, success is determined based on two criteria. Firstly, the picking object's bounding box should intersect with the receptacle trigger box (which is different from the bounding box and only includes the area of the receptacle, \eg internal rectangular area of a pan without the handle) when both objects static. Secondly, the distance between the picking object and the center of the receptacle trigger box must be within a threshold to avoid edge cases or large receptacles. In the case of random targets or point goals, only the second criterion is used.

\noindent \textbf{Objects Partitions.} We illustrate the objects to be picking household objects and target receptacles for each object budget level and unseen tests (\textsc{Obj-OoD}, \textsc{All-OoD}) in Table.~\ref{tab:stretch_obj_partitions}.

\begin{table}[htbp!]
    \centering
    \small
    \begin{tabularx}{\columnwidth}{>{\centering\arraybackslash}m{0.2\columnwidth} >{\raggedright\arraybackslash}m{0.33\columnwidth} >{\raggedright\arraybackslash}m{0.35\columnwidth}}
        \centering
        \textbf{Levels} & \textbf{Picking object} & \textbf{Receptacle} \\ \midrule
        \hline
        Budget $=1$ & red apple & striped plate \\ 
        \hline
        Budget $=2$ & red apple, bread & striped plate, baking pan \\ 
        \hline
        Budget $=4$ & red apple, bread, sponge, blue cube & striped plate, baking pan, metal bowl, saucer \\
        \hline
        Unseen objects & green apple, tomato, potato, mug & wooden bowl, bamboo box, rusty pan, pot \\ \bottomrule
    \end{tabularx}
    \caption{\textbf{Objects partitions for different object budget levels and unseen tests in \bench.}}
    \label{tab:stretch_obj_partitions}
\end{table}

\noindent\textbf{Texture and materials randomizations.} Further, for texture and lightning randomization evaluations, we randomize the RGB color of materials in range $[0,0,0]$ to $[255,255,255]$ and over 5 different materials for the tabletop (table surface and legs are randomized separately), surrounding walls, and floor. The color of the lights are randomized similarly and the intensity is randomly sampled from $[0.5, 2]$, where during training the light intensity is 1.

\noindent\textbf{Unseen room structures.} During the training, there is no extra furniture other than the table in the room, where during the \textsc{All-OoD} evaluation, we added fridges, shelves, flowers, \emph{etc.} as distractors in the background of agent's view. Visualizations can be found in Fig.~\ref{fig:benchmark}

\subsection{Sawyer Peg}\label{appendix:env:sawyer_peg}
\noindent \textbf{Description.}
We use the same Sawyer Peg simulation environment proposed in~\cite{yu2019meta, sharma2021earl}. This task requires the robot arm to grasp the peg and then insert it into a hole of the box. The success criteria is defined by $||s - g|| \leq \epsilon$ where $s, g$ denotes the state and goal, and $\epsilon$ is the small tolerance. We use the distance between the position of the peg and the center of the hole with $\epsilon=0.05$ for all Sawyer Peg experiments same as~\cite{yu2019meta, sharma2021earl}. Action space consists of 3D end-effector delta control and 1D gripper open/close control. For visual task, the visual input includes the wrist-centric camera view and the third-person view with $84\times84$ dimension for each, adapted from~\cite{hsu2022vision}. We also include the 3D end-effector position relative to the robot base, 1D gripper width, and randomly sampled goal as low-dimensional observations. No ground-truth position of the object peg is exposed to the agent.

\noindent \textbf{Training and Evaluation Protocol} Recall Fig.~\ref{fig:sawyer_peg_method_comparison} and Sec.~\ref{sec:sawyer-peg-rgb}. Agents are trained in the usual Sawyer-Peg setting (with one box and hole position) and then tested with both novel hole locations and with novel box positions \emph{and} novel hole positions. Visualizations of the novel box hole can be found at Fig.~\ref{fig:benchmark}, where the hole is bounded with the green square for demonstration purpose, but will not be in agent's visual observations. 

\noindent \textbf{Simulation Stability.}\label{appenidx:env:sp:sim_stabilization}
We found that when the simulation is not reset over a long time horizon, there are unexpected aberrant outcomes in the Sawyer Peg environment\footnote{\href{https://github.com/Farama-Foundation/Metaworld/issues/373}{https://github.com/Farama-Foundation/Metaworld/issues/373}}.  Specifically, after collisions with the gripper or the peg box, the target peg object may no longer be capable of laying flat along the table as expected. To address this issue we regularly reset the position of the peg to a ``flat'' state without otherwise changing the position of the agent's arm or the peg. This reset is different than those used when training episodic agents and is only done to maintain simulator stability when not executing a full environment reset for many timesteps.
We also encountered floating errors and collisions, which are expected and, in some cases, inevitable. Most of these issues occurred on a reasonably small scale (i.e., less than 1e${-}$7) and can be considered as noise. However, some collisions with the gripper and the stationary peg box were more significant and observable. We note that these collisions can lead to (near-)irreversible states which may be challenging, or even impossible, to recover from. The existence of these issues within a simulated environment emphasizes the need for methods, like those proposed in this work, that automatically allow for detecting near-irreversible transitions even when those transitions are impossible to anticipate a priori.
\section{Implementation Details}\label{appendix:impl}

\subsection{Measures of Irreversibility}\label{appendix:measure}

\noindent \textbf{NI Transition.} Here we describe more formally what we mean by a near-irreversible (NI) transition. As usual in reinforcement learning for embodied agents, we formalize our setting as a Partially Observed Markov Decision Process (POMDP) with state space $\cS$, partial observation space $\cO$, action space $\cA$, transition probability measure $P_T$, and reward structure $R$. For simplicity of presentation we will assume that $\cS$ and $\cA$ are discrete. The goal is to learn a policy $\pi$, \ie, a function that maps partial observations to distributions over actions, which maximizes the expected future $\gamma$-discounted rewards ($\gamma\in[0,1]$). We will say that there is a \emph{path} from $s\in\cS$ to $s'\in \cS$ if there exists some $n{\geq}0$, $s_1,...,s_{n-1}\in \cS$, and $a_0,...,a_{n-1}\in\cA$ such that $P_T(s_0, a_0, s_1),..., P_T(s_{n-1}, a_{n-1}, s_n)>0$ with $s_0=s$ and $s_n=s'$. We say that $s,s'$ are connected if there exist paths from $s$ to $s'$ and from $s'$ to $s$. Similarly we will say that a set $S\subset \cS$ is connected if all $s,s'\in A$ are connected. Note that connectedness an equivalence relationship and thus partitions $\cS$ into equivalence classes, we call these classes \emph{connected components}.

We assume that, during training, we never give the agent a goal that requires it to undergo an irreversible transition. More formally, if we let $\cG\subset\cS$ be the set of goal states and $\cI\subset \cS$ be the set of all possible states to which the agent may be reset, then we assume that $\cG{\cup}\cI$ is connected. Finally we are in a place to define what we mean by an irreversible transition. Let $\tau_\pi(i)\in\cS$ for $0\leq i$ be a random variable representing the trajectory of an agent with policy $\pi$ where $\tau_\pi(0)$ represents the start state of the agent after a reset (\ie, $\tau_\pi(0)\in \cI$). Moreover, let $\cU\subset\cS$ be the connected component containing $\cI{\cup}\cG$. Then we say that the agent has undergone an \emph{irreversible transition} at step $i$ if $\tau_\pi(i) \in \cU$ and $\tau_\pi(i+1)\not\in\cU$. We will also call every state $s\in\cS\setminus\cU$ an \emph{irreversible state}. See Fig.~\ref{fig:irreversibility-types} for examples of reversible and irreversible states in \bench.

While this definition of irreversibility reflects the capabilities of the environment, it falls short in two ways. First, during training, we should not care if a state $s$ is truly irreversible, we instead need to consider whether or not the agent's current policy $\pi$ has any hope of ever moving from $s$ to a goal as, otherwise, it cannot hope to succeed. Secondly, with our current definitions, a state $s$ is considered reversible even when the probability of the agent ever reaching a state in $\cI{\cup}\cG$ from $s$ is arbitrarily small. For instance, in Fig.~\ref{fig:irreversibility-types} we see examples of what we call ``near-irreverable'' states where the agent must execute a long sequence of precise steps to retrieve the object. To formalize, we will say that a state $s$ is \emph{$(\pi, \epsilon, N)$ near-irreversible} if, when following $\pi$, the probability of reaching a goal state from $s$ within $N$ steps is $<\epsilon$.

Computing precisely whether or not a state $s$ is $(\pi, \epsilon, N)$ near-irreversible, or even irreversible, is computationally intractable even with full knowledge of the environment. Instead, we use the above formalization to build intuition for the behavior we expect to see when an agent has experienced a near-irreversible transition. 

\noindent \textbf{Measures of reversibility.}
We now introduce the details of our reversibility measures. We propose two domains of measures: dispersion-based and distance-based. Intuitively, the dispersion-based approaches measure the dynamics of the trajectories with metrics like variation/standard deviations or entropy, where the distance-based methods measure measure the distance from recent states to previous states to characterize how far and how often the agent visits new states or whether it is trapped within a few states.

\noindent \textbf{Dispersion-based measures.} We illustrated the dispersion-based method in Algorithm.~\ref{algo:dispersion}. The standard deviation and entropy metrics are used as the baseline for our dispersion-based measurements. To elaborate, we denote the policy $\pi$, dispersion metrics $\phi$, threshold $\epsilon$, trajectory buffer $\tau$, and horizon for measurement check $N$. Then, we simply define the dispersion measure over the trajectory $\tau = \{s_0, s_{1}, \cdots, s_N\}$ sampled from on-policy $\pi$ as $\phi(\tau)$. And once $\phi(\tau)$ consistently smaller than the threshold $\epsilon$, then we say the current measure stage as $(\pi, \epsilon, N)$ is near-irreversible (NI). We directly use $\phi(\tau) = std(\tau)$ for our method \textsc{Std}. And we calculate the entropy over a discretized state space as our \textsc{Ent} method. In practice, to characterize the consistency of near-irreversible behaviors, we introduce the hyperparameters $n_{tol}$ for the number of consecutive phases that the measurement is considered near-irreversible states. In all of our experiments, we use the $N=300$ and $n_{tol} = 2$.

\RestyleAlgo{ruled}
\begin{algorithm}[hbt!]
\SetKwInput{kwInit}{Init}
\caption{Dispersion-Based Measure}\label{algo:dispersion}
\kwInit{policy $\mathcal{\pi}$}
\kwInit{$\phi, \epsilon, \tau, N$, optional $n_{tol}$ and $n_{irr} = 0$} %
\While{not done}{
  
    {\color{darkgreen} \# randomly sample goal from the goal space}\
    sample $g\in\mathcal{G}$\;
    rollouts in environment with $\pi$ and update $\tau$\;
    {\color{darkgreen} \# near-irreversible formulation in Appendix.~\ref{appendix:measure}} \\
    $NI(\pi,\epsilon, N) =$ False\;
  
    \If{len($\tau$) $\geq N$}{
        {\color{darkgreen} \# dispersion mesure} \\
        $\rho = \phi(\tau)$ \;
        \eIf{$\rho < \epsilon$}{
            $n_{irr} += 1$\;
            {\color{darkgreen}\# consecutive near-irreversible check}\\
            \If{$n_{irr} \leq n_{tol}$}{
                $NI(\pi,\epsilon, N) =$ True\;
                $n_{irr} = 0$\;
            }
        }{
            $n_{irr} = 0$\;
        }
        clear $\tau$\;
    }
    update $\pi$ with PPO~\cite{Schulman2017PPO}\;
}
\end{algorithm}

\noindent \textbf{Distance-based measures.}
Similarly as the dispersion-based measures, our distance-based measures are also calculated over the trajectory. We illustrated the distance-based method in Algorithm.~\ref{algo:distance}. We propose \textsc{L2} for euclidean distance metric, and \textsc{Dtw} for dynamic time wrapping metric as $d$ in the algorithm. The measure is calculated as the distance from a sliding window of the trajectory with total length $M$, to the previous histories, which is also a sliding window with the maximum length $N-2M$. Besides the traditional euclidean distance metrics, our setting is intuitively and naturally allows for using a DTW measures as the distance measure $M$, trajectory horizon $N$, and sliding windows have different length, makes the DTW efficiently measure the distance between states. We use $M=100, N=600$ (which shares the same history length as dispersion-based method but without the consecutive phases check) for all experiments in all simulations. 

\RestyleAlgo{ruled}
\begin{algorithm}[hbt!]
\SetKwInput{kwInit}{Init}
\caption{Distance-Based Measure}\label{algo:distance}
\kwInit{policy $\mathcal{\pi}$}
\kwInit{$d, \epsilon, \tau, N$, measure steps $M < N/2$} %
\While{not done}{
  
    {\color{darkgreen} \# randomly sample goal from the goal space}\\
    sample $g\in\mathcal{G}$\;
    rollouts in environment with $\pi$ and update $\tau$\;
    {\color{darkgreen} \# near-irreversible formulation in Appendix.~\ref{appendix:measure}}\\
    $NI(\pi,\epsilon, N) =$ False\;
  
    \If{len($\tau$) $\geq N$}{
        {\color{darkgreen} \# distance from slide-window to past}\\
        distances $= []$ \;
        \For{$i\gets0$ \KwTo $M$}{
            $s = \tau[N- M + i]$\;
            $past = \tau[:N - 2M + i]$\;
            $d_{min} = \min(d(s, past))$\;
            Add $d_{min}$ \KwTo distances\;
        }
        $d_{maxmin} = \max$(distances)\;
        \If{$d_{maxmin}  < \epsilon$}{
            $NI(\pi,\epsilon, N) =$ True\;
        }
        clear $\tau$\;
    }
    update $\pi$ with PPO~\cite{Schulman2017PPO}\;
}
\end{algorithm}

\subsection{Irreversibility Threshold}\label{appendix:sawyer_peg_threshold}

\begin{figure*}[htbp!]
    \centering
    \includegraphics[width=\linewidth]{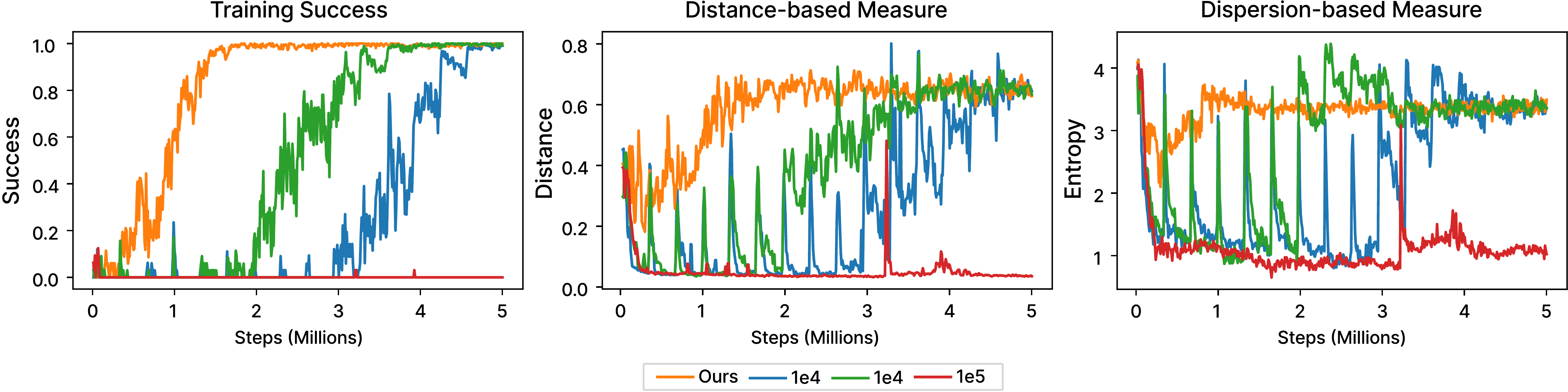}
    \caption{\textbf{Visualizing reversibility measures during Sawyer-Peg training.} Notice that the periodic methods (1e4-green, 1e4-blue, and 1e5-red) methods have low reversibility values during the majority of training (right two figures) with periodic jumps in these values. These periodic jumps correspond to resets and suggest that these agents spend a large portion of their training time in near-irreversible states waiting for a reset to begin learning again. Unlike these methods, our baseline has a consistently high reversibility values across training (as the agent will reset when such values become too low) and learns quickly. For demonstration purpose, all models here are trained using 32 parallel processes.
    }
    \label{fig:measure_thresh}
\end{figure*}

In Fig.~\ref{fig:measure_thresh}, green and blue curves used the same periodic reset setting (resetting every 1e4) steps but, due to randomness in RL training, converge at different timesteps. The orange curve (ours) learns faster at the beginning of the training and the red curve (1e5 steps/reset) learns little in 5M steps. We found that both our dispersion-based and distance-based measures effective characterize the learning process and capture those timesteps where periodic resets occur (measures of reversibility become large after resets). In particular, those increases in the reversibility measures capture that the agent can attempt to reach the peg after a reset but, but (as shown the the gradual decline of the reversibility measures) unfortunately pushes the peg into a position that is difficult to reach due to the sub-optimal policy early in training. Selecting a threshold (recall the $\alpha$ parameter from Sec.~\ref{sec:quantifying-irr} of the main paper) for deciding when to reset based on the reversibility measure is trivial: we select it as some value between the those values obtained by a random policy soon after a reset and those values obtained by doing nothing. Empirically we found that reasonable thresholds are also robust and can be adjusted to fit one's preferences for reset frequency: with a smaller value, we may expect the agent to require fewer resets but a larger number of total training steps to solve the task, and vice-versa. The horizon for irreversible check is even more robust in general. We consistently use 600 steps across all experiments (which corresponds to twice the maximum number of steps in a training episode). Meanwhile, instead of learning being hampered primarily by the challenge of exploration (as argued in~\cite{sharma2021earl}), we claim that the main challenge for autonomous RL is actually the existence of near-irreversible states: the agent is able to explore but only so long as it hasn't entered such a state.

\subsection{Algorithms}\label{appendix:algo}

To clearly distinguish between different training methods for traditional episodic or autonomous RL, we now provide pseudocode for the reset function in episodic RL (Pseudocode.~\ref{supp:code:episodic_reset}), our single policy with measurement-determined resets with random goals (Pseudocode.~\ref{supp:code:single_policy_reset}), and previous FB-RL with periodic and supervised explicit irreversible interventions (Psudocode.~\ref{supp:code:FB_policy_reset}). In our experiments, for the FB-RL{$+$}GT baseline, we simply replace irreversibility prediction network proposed in~\cite{xie2022ask} with ground-truth values, this should represent an upper bound on the performance of this method. Nevertheless, as described in previous sections, we found that the FB-RL{$+$}GT method underperformed our proposed approach, likely due to the existence of near-irreversible states that cannot be easily captured as GT labels.

\begin{minipage}{0.9\linewidth}
\begin{lstlisting}[language=Python,label={supp:code:episodic_reset},caption={Reset function in episodic setting}]
def episodic_reset():
    reset_environment(
        agent_initial_state, 
        arm_initial_state, 
        object_initial_states
    )
    # similar start every episode
    agent_start = agent_initial_state
    arm_start = arm_initail_state
    object_start = object_initial_states
    goal = target
    return get_obs()
\end{lstlisting}
\end{minipage}

\begin{minipage}{0.9\linewidth}
\begin{lstlisting}[language=Python,label={supp:code:single_policy_reset},caption={Reset function in our random targets with measurement-lead intervention setting}]
def reset():
    if measure_check_near_irreversible():
        # measurement-led intervention
        return episodic_reset()
    else:
        # continued from previous phase
        agent_start = agent_prev_state
        arm_start = arm_prev_state
        object_start = object_prev_state
        # any state exclude object state
        # prevent success immediately
        goal = random.sample(
            goal_space \ object_start
        )
        return get_obs()
\end{lstlisting}
\end{minipage}

\begin{minipage}{0.9\linewidth}
\begin{lstlisting}[language=Python,label={supp:code:FB_policy_reset},caption={Reset function in previous FB-RL with periodic intervention setting.}]
def reset():
    # switch phases
    phase = 1 - phase
    if cumulative_steps >= reset_frequency:
        # periodic intervention
        return episodic_reset()
    elif is_explicit_irreversible():
        # Either supervised or ground truth
        # by `object_height < table_height`
        return episodic_reset()
    else:
        # continued from previous phase
        agent_start = agent_prev_state
        arm_start = arm_prev_state
        object_start = object_prev_state
        if phase == 0:
            # forwad phase
            goal = target
        else:
            # backward phase
            goal = random.sample(
                initial_states
            )
        return get_obs()
\end{lstlisting}
\end{minipage}

With goal-conditioned POMDP defined in Sec.~\ref{sec:single-policy}, in FB-RL, in FB-RL setting, the forward goal space is normally defined as the singleton $\mathcal{G}_{f} = \{g^\star\}$ for the target task goal $g^\star$ (\eg the apple is on the plate, the peg is inserted into the hole, etc), where the goal space for backward phase is exactly the (limited set of) initial state space $\mathcal{G}_{b} = \cI \subset \cS$. Empirically, those initial states are pre-defined and discriminated in previous work such that the deployed $\mathcal{G}_f$ and $\mathcal{G}_b$ are disjoint. Then it is reasonable to use separate policies and objectives to optimize alternatively during the training. However, as in our setting, by simply given $\mathcal{G} = \mathcal{S} \setminus \{s_t\mid s_t\in\tau_{\pi} (t) \in \cT_\pi, \Phi_{W,N,\alpha,d, \pi} (\cT_\pi) = 1\}$ which is represented as the entire state space excluding the NI states we formalized above given the current $\pi$ during training, we can make the analogue of each \textit{phase} in our autonomous RM-RL setting and using the same objective as an episode noted in episodic RL, where by denoting a \textit{phase$_i$} with states trajectory $\{s^i_0, \cdots, s^i_H\}$ as an $i^{th}$ episode with state $s^i_t \in \cS$ at timestep $t$, finite horizon $H_i$, and goal $g_i\in \mathcal{G}$ such that $s^i_H=s^{i+1}_0$, \ie the last state of the $i^{th}$ phase is identical to the initial state of the $(i+1)^{th}$ phase. Therefore, we can directly use the same objective with switching goals without a reset, where a reset thereby can be simply defined as an intervention such that $s^i_H\neq s^{i+1}_0$. Empirically, we define a reset or an intervention in embodied tasks at the beginning of the episode or phase, entirely or partially from 1) teleport the agent, 2) teleport the arm, and/or 3) put objects in environment to some initial configurations. Unlike some previous reset-free work, we still want to highlight that even a hard-code teleportation without human intervention in real world or simulation is also restrictively counted as a reset in our work.
 
Except the goal updates, which can be done automatically, all other resets require human intervention (including the hard-coded programming for agent or arm teleportation in the real world or simulation) in practice. As shown in Pseudocode.~\ref{supp:code:single_policy_reset}, by eliminating the ``reset'' gameplay from prior work, our method is similar to the episodic setting except that the start state of an phase is exactly the same as the last state of previous phase. Note that the goal space for each phase is symmetric by sampling randomly (exclude the state that the object is exactly at the goal, \ie succeed immediately)

\subsection{Model Architecture}\label{appendix:arch}

\noindent \textbf{Baseline Model for \bench}
As show in Fig.~\ref{fig:architecture}, our baseline model conditioned on visual observations, language instruction, and low-dimensional proprioception and goal state. Specifically, we used the frozen CLIP ResNet50 as the visual backbone followed with a CNN compressor consisting two convolutional layers with 128 and 32 channels respectively. Each convolutional layer had a kernel size of $1\times1$ and was activated with ReLU. In addition, we discarded the final average pooling and linear layers from the CLIP ResNet50, and only kept the last spatial map before the trainable CNN compressor. And we replace the batch norm with group norm. For the language stream, tokenized texts are passed to the frozen CLIP language encoder and projected with one 1024 linear layer and expanded as the same dimension of the output spatial map like~\cite{shridhar2022cliport}, and passed to a CNN combiner with the output of the CNN compression from the visual stream. The CNN combiner has the same channels and kernels as the compressor. Finally, for the low-dimensional point observation, we simply encode it with a linear layer followed with the layer normalization and fuse it with the vision and language embedding and then pass to the actor and critic for on-policy updates. The actor and critic are defined as linear layers with output dimensions of 10 and 1 respectively.

\noindent \textbf{Sawyer Peg}
For a fair comparison, we modify our model in Fig.~\ref{fig:architecture} by replacing the pretrained vision and language backbone with the CNN backbone from scratch and with random crop and shift augmentations like ~\cite{hsu2022vision, Sharma2023SelfImproving}. However, unlike ~\cite{hsu2022vision, Sharma2023SelfImproving} which use separate vision encoders, we use only \textit{single} CNN encoder (just like our idea of single policy) that digest both views of image for parameter-efficiency. We ended with finding that single visual encoder is enough for solving this task. The low-dimensional observation is encoded as the same way as for \bench. The output is then fused with the visual representation and passed into actor and critic networks with two hidden layers of 512 and 256 unit each. We further follow the practice from~\cite{tang2020discretizing,brohan2022rt} that discretizes the continuous action space into multi-discrete action space with 7 bins of each dimension (\ie $4\times7=28$ in total) and use multi-categorical distribution for our policy. We find slightly better empirically in this task and efficient than using parameterized scale for Gaussian distribution.

\noindent \textbf{ObjectNav}
Same as our baseline model but without the frozen language backbone and include extra GRU state encoder for consistent comparisons with previous work \cite{Khandelwal2022SimpleButEffective}.

\subsection{Reward Shaping}

Instead of costly process of collecting demonstrations (even from the oracle agent trained in episodic settings), for \textit{all} autonomous training in \bench, Sawyer Peg, and ObjectNav, we adopt a \textit{unified} distance-difference-based reward $R$ given state transition $s_{t-1}, s_t \in \mathcal{S}$ and goal $g\in\mathcal{G}$ as:
\begin{equation*}
    R(s_t | g) := \alpha\cdot(d(s_{t-1}, g) - d(s_t, g)) + \beta \cdot \mathbbm{1}_{success}
\end{equation*}
where $d$ is the non-negative distance function, $\mathbbm{1}_{success}$ is the indicator function for the task success, and $\alpha, \beta$ are weighted multipliers. We simply set $\alpha=1, \beta=10$ for experiments except in Sawyer Peg where $\alpha=100$ due to its smaller state space. The first term incentives the agent to continuously make progress towards the goal while the second terms is the terminal reward for completing the task. This reward formalism is first proposed in \cite{ng1999policy} and widely used in navigation tasks. Recently it has also been used as an implicit reward for goal-conditioned pretraining in embedding space~\cite{lee2021generalizable,li2022phasic,ma2022vip}. In practice, we calculate the Euclidean distance from the gripper to the object and from the object to the target for manipulation experiments, and the distance from the agent to the goal object for object navigation experiments. Note that neither the object positions nor the agent base GPS positions is included in the observations during the training and inference. We believe that a \textit{single} reward shaping that can be easily applied to tasks in different embodied space does not require tedious engineering, and it is applicable and practical for the general formulation of the autonomous RL.

\subsection{Training Hyperparameters}\label{appendix:training_details}

To train the policy with RL, we use PPO with Generalized Advantage Estimation (GAE) and normalized advantages~\cite{schulman2015gae}. For paralleling training, we adopt DD-PPO in AllenAct~\cite{Weihs2020Allenact} with 1 worker on each GPU. We detail the shared (left) and different (right) default hyperparameters for training in each task domain.

\begin{table*}[htbp!]
\centering

\begin{minipage}{.3\linewidth}
    \centering
    \begin{tabular}{@{}cc@{}}
        \toprule
        \textbf{Hyperparameter} & \textbf{Value} \\ \midrule
        GPU instance & \texttt{g4dn.12xlarge} \\
        Worker per GPU & 1 \\
        Optimizer & Adam~\cite{kingma2014adam} \\
        Discount factor $\gamma$ & 0.99 \\
        GAE $\tau$ & 0.95 \\
        Value loss coefficient & 0.5 \\
        Normalized advantages & True \\
        Max gradient norm & 0.5 \\
        Entropy coefficient & 0.01 \\ \bottomrule
    \end{tabular}
\end{minipage}
\hfill
\begin{minipage}{.6\linewidth}
    \centering
    \begin{tabular}{@{}lccc@{}}
        \toprule
        \textbf{Hyperparameter/Value} & \textbf{\bench} & \textbf{Sawyer Peg} & \textbf{ObjectNav} \\ \midrule
        Number of GPUs & 2 & 2 & 4 \\
        Environments per GPU & 8 & 4 & 15 \\
        Learning rate & 3e-4 & 5e-4 & 3e-4 \\
        Rollout length & 200 & 1024 & 128 \\
        PPO epochs & 10 & 20 & 4 \\
        Number of mini-batches & 4 & 1 & 1 \\
        PPO Clip & 0.1 & 0.2 & 0.1 \\\bottomrule
    \end{tabular}
\end{minipage}

\caption{Shared (left) and separate (right) hyperparameters for \bench, Sawyer Peg, and ObjectNav experiments}
\label{tab:train_param}
\end{table*}

\section{Initial Results in Reset-Minimization for Object Navigation}\label{sec:objectnav}

\begin{figure*}[t]
    \centering
    \includegraphics[width=\linewidth]{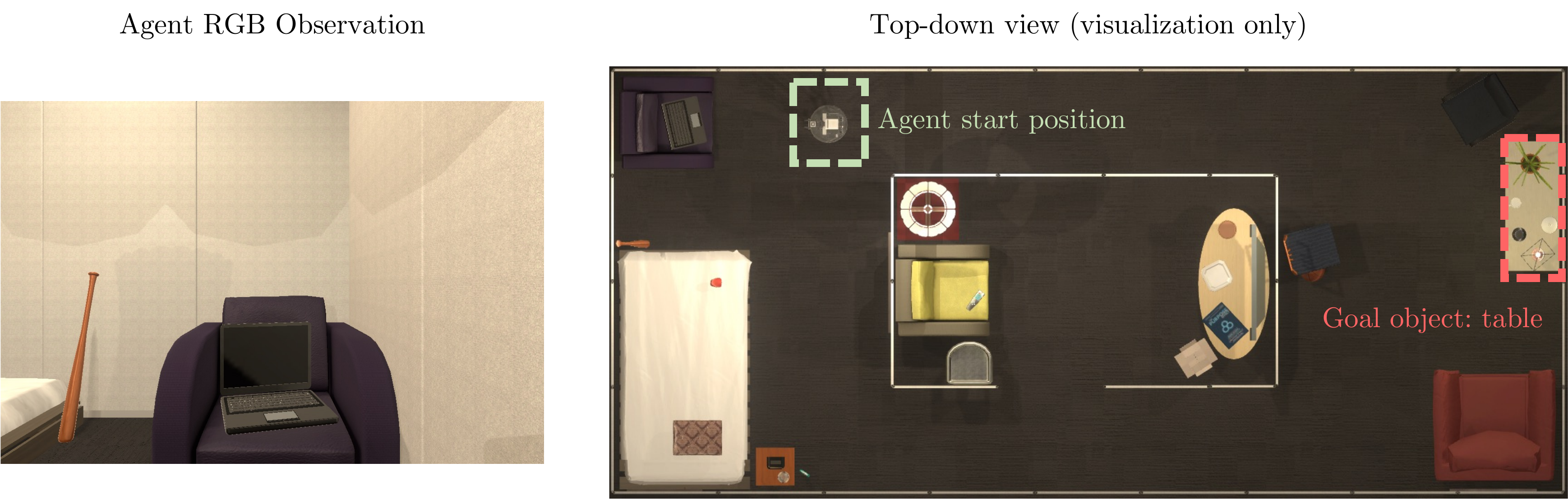}
    \caption{\textbf{RoboTHOR Object Navigation Task.} An example of a the start of a RoboTHOR ObjectNav episode. The agent is given an RGB observation (left) and must navigate to an object of a goal category (table). On the right we show a top down map view of the environment with the agent's position and goal object highlighted. Note that this top down view is for visualization only and is not available to the agent at training or inference time.}
    \label{fig:robothor-objectnav}
\end{figure*}

To highlight the general applicability of our proposed RM-RL approach for embodied AI tasks, we show preliminary results on the RoboTHOR Object Navigation (ObjectNav) task~\cite{Deitke2020RoboTHORAO}. In ObjectNav, an agent is placed within a simulated household environment and given a goal object category (\ie TV, chair, table, \etc), see Fig.~\ref{fig:robothor-objectnav}. The agent must then navigate to an object of this category using only visual observations. The agent is considered to have successfully completed the task if it executes a special \textsc{End} action and an object of the given category is both visible and within 1m of the agent. Regardless of whether or not the success criteria are met, the \textsc{End} action always ends the episode. Note that, in principle, there are no irreversible states in ObjectNav: the task merely involves navigating around an, otherwise static, environment. Nevertheless, as we show below, choosing when to reset carefully can result in significant improvements in efficiency. 

\subsection{Training and Evaluation Protocol}

We extend the RoboTHOR ObjectNav training code available in AllenAct~\cite{Weihs2020Allenact}. This training code contains the EmbCLIP model introduced by Khandelwal \emph{et al.}~\cite{Khandelwal2022SimpleButEffective} which is SoTA among models trained purely upon the RoboTHOR ObjectNav dataset (there exist more performant models, \eg those pretrained upon ProcTHOR~\cite{Deitke2022ProcTHOR}, but these use extensive additional pretraining data in the form of procedurally generated environments). The training set contains 60 different house plans with 12 goal object types. We use 60 separate training processes for all experiments.

In order to train in a reset-minimizing setting, we create a variant of the RoboTHOR ObjectNav task where the agent is not automatically reset in the usual episodic fashion. In particular, we turn the \textsc{End} into a soft version where, if the agent executes the \textsc{End} action without success criteria being met, then the agent is penalized and the goal object remains the same. If success criteria are met or reach 300 steps for the current goal, however, then the agent is given a new goal object category within the same environment. As usual, we allow the agent to request a full environment reset and we use our usual measures of irreversibility to determine when these resets should be executed. We call training variant RM-ObjectNav. 

All models are evaluated on the RoboTHOR validation set as usual, \ie with an \textsc{End} action that immediately ends the episode regardless of success. In particular, the agent is evaluated with 1800 tasks from 15 unseen houses but with the same target object types as training.

\subsection{Results}

We show our training results in Fig.~\ref{fig:objnav_train} and evaluation results in Table~\ref{tab:objnav_result}. In particular, we show validation set performance after 50M and 100M training steps. Our baselines include four models trained in the RM-ObjectNav setting: (Ours) an EmbCLIP baseline agent with resets taken when using our \textsc{Std} irreversibility measure and ($N=300$, $N=10k$, $N=\infty$) three models with periodic resets taken after every 300, 10k, and $\infty$ steps. We also include a~\cite{Khandelwal2022SimpleButEffective} baseline trained in the usual ObjectNav setting (\ie without the ``soft'' \textsc{End} action). To emphasize: all models are evaluated with hard \textsc{End} actions.

The results from Table~\ref{tab:objnav_result} are very promising: after 100M steps with only 635 resets we are able to achieve success rates \textbf{higher} than all competing baselines despite the next best performing baseline using 2M resets. Note also that our agent takes the vast majority (592 of 635) of its resets within the first 50M steps of training showing that our model continues to learn (going from a success rate of 0.216 at 50M training steps to 0.551 at 100M training steps) using very few resets. Recall that, technically, there are no explicitly irreversible states in ObjectNav. Nevertheless, we show that pure Reset-Free training results in slower training convergence and lower evaluation results. In comparison with other periodic reset competitors trained with the same soft \textsc{End} action, we show that all of them provide lower evaluation results than ours and EmbClip, mainly due to the false positive that early stopped during the evaluations. Therefore, we believe providing a very few resets at proper time-point determined by our unsupervised method will give more efficient performance in general.

\end{document}